\documentclass{article}

\usepackage{arxiv}

\usepackage[utf8]{inputenc} 
\usepackage[T1]{fontenc}    
\usepackage{hyperref}       
\usepackage{url}            
\usepackage{booktabs}       
\usepackage{amsfonts}       
\usepackage{nicefrac}       
\usepackage{microtype}      
\usepackage{lipsum}
\usepackage{graphicx}
\usepackage{times}
\usepackage{epsfig}
\usepackage{amsmath}
\usepackage{amssymb}
\usepackage{float}
\usepackage{multirow}
\usepackage{subfigure}
\usepackage{setspace}
\graphicspath{ {./images/} }

\title{Association: Remind Your GAN not to Forget}

\author{
 Yi Gu$^{\dagger}$  Jie Li$^{\dagger}$ $^{\ast}$  Yuting Gao$^{\ddag}$  Ruoxin Chen$^{\dagger}$  Chentao Wu$^{\dagger}$  Feiyang Cai$^{\S}$  Chao Wang$^{\dagger}$  Zirui Zhang$^{\dagger}$ \\
 $^{\dagger}$ Department of Computer Science and Engineering, Shanghai Jiao Tong University, Shanghai, China\\
 $^{\ddag}$ Department of Electrical and Computer Engineering, Texas A\&M University, College Station, US\\
 $^{\S}$ Department of Medicine, McGill University, Montreal, Canada \\
 \texttt{\{louisgu,lijiecs\}@sjtu.edu.cn}
}

\begin{document}
\maketitle

\begin{abstract}
Neural networks are susceptible to catastrophic forgetting. They fail to preserve previously acquired knowledge when adapting to new tasks. Inspired by human associative memory system, we propose a brain-like approach that imitates the associative learning process to achieve continual learning. We design a heuristics mechanism to potentiatively stimulate the model, which guides the model to recall the historical episodes based on the current circumstance and obtained association experience. Besides, a distillation measure is added to depressively alter the efficacy of synaptic transmission, which dampens the feature reconstruction learning for new task. The framework is mediated by potentiation and depression stimulation that play opposing roles in directing synaptic and behavioral plasticity. It requires no access to the original data and is more similar to human cognitive process. Experiments demonstrate the effectiveness of our method in alleviating catastrophic forgetting on image-to-image translation tasks.
\end{abstract}

\begin{figure}[H]
		\hsize=\textwidth 
		\centering
		\includegraphics[scale=0.47]{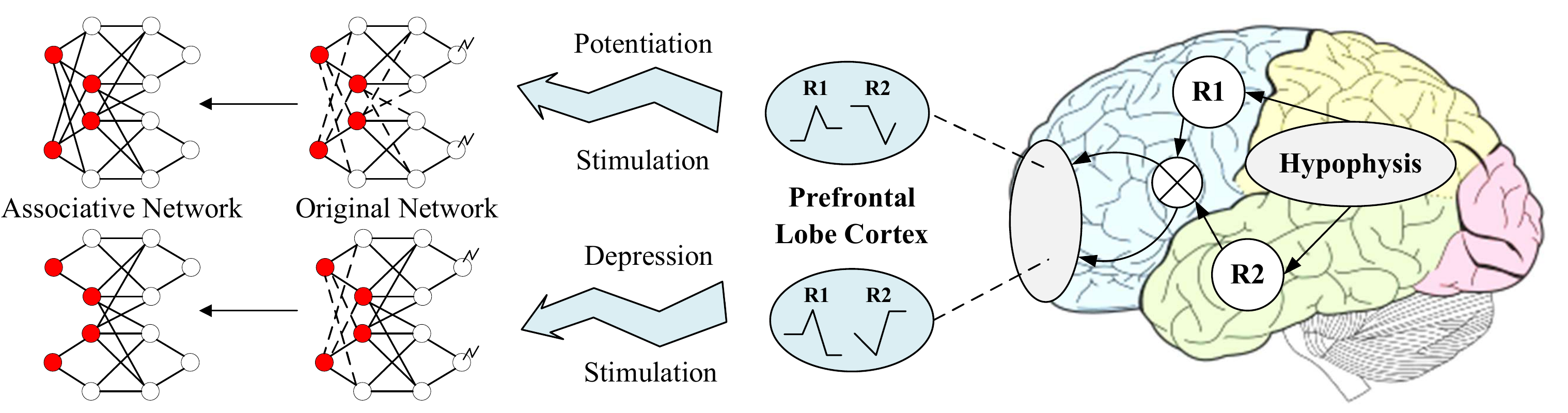}
		\caption{The formation of association depends on the dopaminergic reinforcement mediated by two dopamine receptors, DopR1 and DopR2. The backward pairing of DopR1 and DopR2 ensures the potentiation while the forward pairing of DopR1 and DopR2 induces the depression. These two signaling cascades exhibit different temporal sensitivity, instructing opposing influences on associative learning. } 
		\label{brain}
\end{figure}

\section{Introduction} \label{introduction}
When tasks or datasets are sequentially and separately fed into the neural network for training, the model usually inevitably encounters the phenomenon of catastrophic forgetting \cite{mccloskey1989catastrophic}. The network constantly forgets the knowledge obtained from previous tasks whilst learning new training samples. This results in an arbitrary degradation in the model performance on historical tasks. Continual learning or lifelong learning is still a long-standing open problem.

Although some efforts \cite{lee2020continual,xiang2019incremental,gaurav2020simple,10.1007/978-3-030-58621-8_23,cong2020gan} have been devoted to tackle the catastrophe forgetting of discriminative models, image generation models, like most image classification ones, still consider that all data is available. It remains an under-explored area for generative models to acquire the continual learning capability, especially for image-to-image translation task. Different from the discriminative task, the continual learning for generative task has intrinsic characteristics. Each training input for the discriminative model has a fixed corresponding label and the output is discrete. However, the generative model output is coherent, which requires the model to maintain the visual perception, i.e., the prediction should be semantically plausible and visually satisfactory. Therefore, the continual learning for paired image generation task can be formulated as enabling the neural network to perceive the reconstruction to new domains while memorizing captured representations.

Memory replay \cite{wu2018memory,lesort2019generative,ye2020learning,zhai2020piggyback} is one common practice to alleviate catastrophic forgetting for generative models. It saves original images or extracted features from historical tasks, and revisits them when the network trains new task. However, this approach results in additional spatial space to store samples and more time consumption for training. Regularization \cite{seff2017continual,liang2018generative,zhai2019lifelong,schmidt2020towards,thanh2020catastrophic} is another countermeasure that has been applied to generative models. It is characterized by adding extra constraints to consolidate acquired knowledge on old tasks without additional storage. However, tasks drawn from different distributions may cause degradation of model performance \cite{wu2018memory}. In essence, learning is a lifelong process for human, the mode that we learn new knowledge is not mechanically memorizing data or models like the above-mentioned methods. Human tend to incorporate the acquired experience, and summarize the correlation between two successively happening events. This memory and cognition mechanism is association. 

Human leverage connections between relative things to associate new memory with recalled past memories, which makes the new knowledge learning more efficient and versatile. This ability is critically determined by neuron network in association cortex. As shown in Figure \ref{brain}, when human brain receives a stimulus from a certain cortex, the screen of events related to the stimulus will appear in the cortex. In the associative learning mode, some neurons are repeatedly stimulated, altering the efficacy of the synaptic transmission. We call this phenomenon synaptic plasticity, which facilitates or represses synaptic transmission by regulating the postsynaptic potential (PSP). If the stimulation is long enough, such PSP will have long-term alteration, including potentiation and depression \cite{xin2020myelin}. Hence, a sensory stimulation will activate relative neurons, and other irresponsible neurons will be activated by these neurons leading to association.

Motivated by this biological mechanism, we propose to imitate the human associative learning to achieve the continual learning for image generation. We design a heuristics module to produce potentiation stimulations for the generative model. Once the signal is released to the model, it triggers synaptic plasticity to alter the efficacy of synaptic transmission, which associate the current screen with historical representation. However, the heuristics module only observes the single side of reconstruction to past domains. Similar to the depression influence in the brain, we distill past knowledge to new domains based on knowledge distillation \cite{hinton2015distilling}. Note that since we do not preserve raw data, we develop a distillation algorithm that supports continual learning from the probabilistic perspective, which adapts the associated feature representation to the updated feature space. Our brain-like associative learning model, Assoc-GAN, closely resembles human learning and cognition process in achieving continual learning.

Contributions in this paper include: 1) We design a general brain-like continual learning framework for generation tasks. 2) A potentiation module that can heuristically associate the current scene with the historical representation without memorizing task data. 3) A continual distillation algorithm that can be applied to a sequence of tasks. 4) Extensive experiments demonstrate that our method alleviates catastrophic forgetting with considerable time and storage overhead, and outperforms recent state-of-the-art methods.

\section{Related Work} \label{relatedwork}
\subsection{Associative Learning}
Associative learning is classified into classical conditioning and operant conditioning. Classical conditioning refers to learning associations between a pair of stimulations, while operant conditioning refers to learning between behaviors and consequences \cite{morris1988neuroscience}. Associative learning forms non-declarative memory or implicit memory, which is developed unconsciously and is independent of medial temporal lobe and hippocampus \cite{dudai1989neurobiology}. 

A reputed model is introduced to explain associative learning \cite{zhang2019robust}. This work regards association cortex as a network of millions of neurons. Each neuron receives input spikes from thousands of other neurons. The distributed region of the association cortex is a set of brain regions that comprises extensive portions of the frontal and posterior midline and inferior parietal lobule \cite{raichle2001default}. Prefrontal cortex is the top-level of association cortex. It regulates associative learning and divergent thinking, which contributes to the abstract reasoning ability. A study \cite{luria2012higher} shows that prefrontal cortex damage impaired association, and association cortex, especially prefrontal cortex, enables associative learning.

Recent studies focus on revealing the function of specific neuron and molecule in association cortex network. Typically, Cho et al. \cite{cho2020cross} find that $\gamma$-frequency synchrony between prefrontal parvalbumin interneurons is indispensable in associative learning. Handler et al. \cite{handler2019distinct} reveal that dopamine-receptor signaling flexibly modulates the associations in a dynamic environment. Moreover, Mukherjee et al. \cite{mukherjee2018infralimbic} show that prelimbic and infralimbic cortex connectivity is essential for associative learning. However, restricted by imaging techniques, researchers cannot restore the distribution of every neuron or analyze the association cortex network precisely, but they find some features of the network as presented above.

\subsection{Continual Learning}
In general, the methods to mitigate the catastrophic forgetting can be grouped into three categories: dynamic architecture, regularization, and memory replay.

Dynamic Architecture. For discriminative tasks, it assigns individual sub-networks for different tasks. PNN \cite{rusu2016progressive} freezes previous networks and creates new ones for future tasks, which uses lateral connections to transfer information across tasks. DEN \cite{yoon2017lifelong} performs selective retraining and dynamically assigns the capacity by adding new neurons. Motivated by GWR \cite{marsland2002self}, Parisi et al \cite{parisi2017lifelong} propose a set of prediction-driven hierarchical self-organizing neural networks. DGM \cite{ostapenko2019learning} uses a dynamic network expansion to ensure sufficient capacity for new tasks. 

Regularization. LwF \cite{li2017learning} encourages the predictions of previous learned network and the current one to be similar. EWC \cite{kirkpatrick2017overcoming} preserves the information of previous task in a posterior. IMM \cite{lee2017overcoming} introduces the mean and mode method to merge the parameters from old and new tasks. IS \cite{zenke2017continual} estimates weights online compared to EWC computes offline. \cite{wiewel2019localizing} introduces a method to specify the catastrophic forgetting of each parameter in the network. \cite{gaurav2020simple} computes the upper bound of absolute forgetting which have a higher variance. \cite{lee2019overcoming,zhai2019lifelong,park2019continual} keep the balance between remembering old tasks and learning new tasks by knowledge distillation \cite{hinton2015distilling}. In terms of generation tasks, Liang et al.\cite{liang2018generative} shows that the training process of GAN is an incremental learning question, in which different cases of the generator in time sequences are treated as individual tasks. LiSS \cite{schmidt2020towards} is a continual learning framework to improve the performance and the robustness of models on the past tasks by using self-supervised auxiliary tasks. Seff et al. \cite{seff2017continual} propose an augmented loss to prevent important weights from suffering huge changes in value. Thanh et al. \cite{thanh2020catastrophic} show that how catastrophic forgetting prevents the discriminator from making the real data point a local maximum.

Memory Replay. iCaRL \cite{rebuffi2017icarl} learns a feature vector and nearest-mean-of-examplars classifier simultaneously. GEM \cite{lopez2017gradient} episodic memory to maintain a subset of learned task. \cite{rostami2020generative,aljundi2019online,xiang2019incremental} store some samples from historical tasks. Continual learning for image generation has achieved great success since the advances of MeRGAN \cite{wu2018memory}. It relies on joint training with replay alignment for pure unpaired image generation. Representation in \cite{xiang2019incremental} consists of the pseudo-sample from old task produced by generator and new sample, and it proves effective in the classification task. \cite{abati2020conditional} is mask-based method and each layer dynamically generate mask given the input. Lesort et al.\cite{lesort2019generative} evaluate and compare generative models by using various strategies to find which generative model is the fittest for incremental learning. LVAEGAN \cite{ye2020learning} not only induces a powerful generative replay network but also learns useful latent representations. 

\section{Brain-inspired Associative Learning}
\subsection{Preliminary}
\begin{figure*}
	\centering
	\includegraphics[scale=0.38]{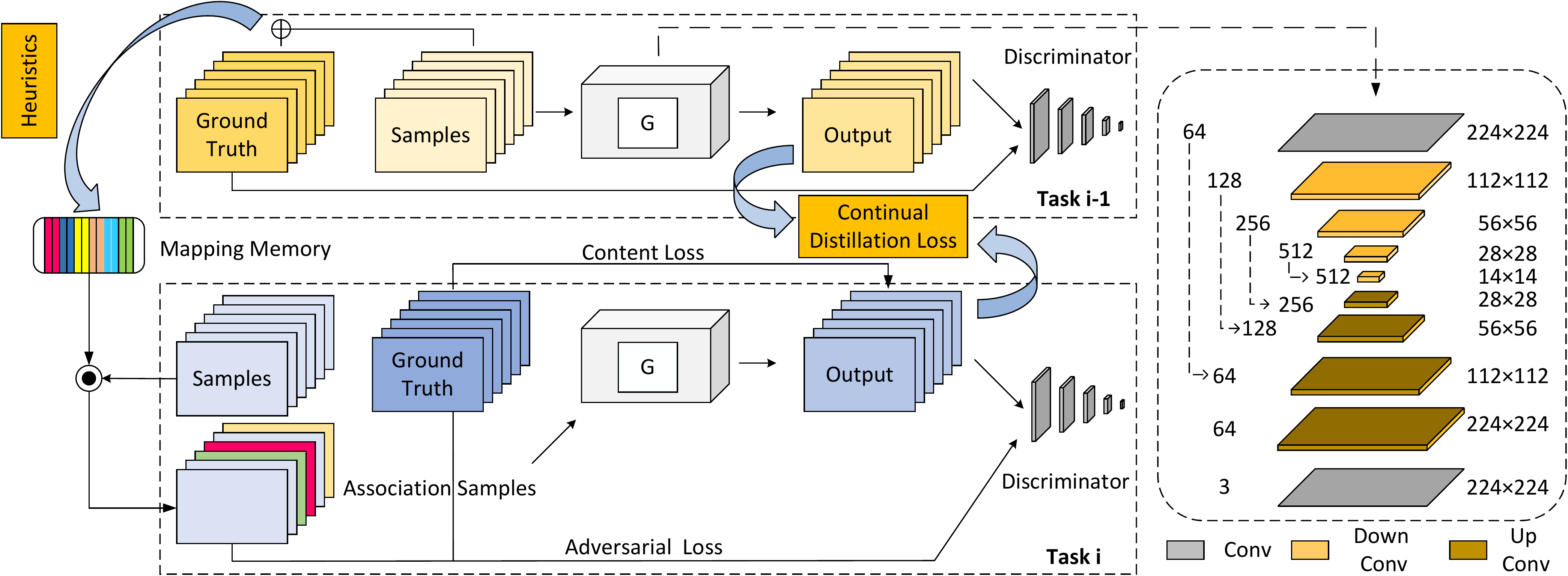}
	\caption{Overview of Assoc-GAN. For task $i$, the model learns to adapt to new environment using given current training samples. To handle the catastrophic forgetting, a heuristics module is adopted to potentiatively associate the current domain with the historical one, and a continual distillation is added to depressively slow down the update of knowledge to the new feature space. }
	\label{framework}
\end{figure*}

Generative network aims to learn a non-linear transformation from source space to target space, which requires pixel-wise supervision. Training is an adversarial process introduced by \cite{goodfellow2014generative}, where the generator $G$ competes with the discriminator $D$ and the two networks are optimized alternately. $G$ aims to generate content that can deceive $D$ while the goal of $D$ is to distinguish between generated samples and real ones. The game can be defined as a minimax objective function: 

\begin{small}
	\begin{equation}
		\max \limits_{\theta_{D}} \min \limits_{\theta_{G}} \mathbb{E}_{z \sim \mathbb{P}_{g}}[1 -\log(D(G(z; \theta_{G}); \theta_{D})] +\mathbb{E}_{\tilde{z} \sim \mathbb{P}_{d} }[\log(D(\tilde{z}; \theta_{D}))].   
		\label{gan}
	\end{equation}
\end{small}where $\theta_{D}$ denotes the parameter of $D$, $\theta_{G}$ denotes the parameter of $G$, $\tilde{z}$ is the target sample subjecting to the real data distribution $\mathbb{P}_{d}$ while $z$ is the source sample subjecting to the mimicked data distribution $\mathbb{P}_{g}$. Here we follow the architectural guidelines of U-Net \cite{ronneberger2015u} as basic generator, which proves the efficacy and simplicity in the pixel-wise prediction of the semantic segmentation, as shown in Figure \ref{framework}. Note that we modify the padding scheme of each convolution layer to keep the input image and output image the same size. To ensure that the prediction is visually realistic and perceptually coherent, we use the discriminative network to distinguish generated images from original ones. Here, the architecture of discriminator is the same as \cite{ledig2017photo} to maintain the generated content to be consistent with the surrounding contexts.

We adopt the content and adversarial loss from some GAN approaches \cite{bai2018finding} \cite{isola2017image} to optimize the generator network. Content loss is a classic choice for measuring perceptual similarity between generated images and original ones. Adversarial loss is adopted to encourage the generator to produce the natural-looking images that manage to fool the discriminator. Therefore, the overall loss function can be formulated as:

\begin{small}
	\begin{equation}
		\mathcal{L}=\mathcal{L}_{mse}+\lambda \mathcal{L}_{adv}  =\sum_{i=1}^{N} \left \|\tilde{z}^{(i)}  -G(z^{(i)}; \theta_{G})  \right \|_{2}    -  \lambda \sum_{i=1}^{N} \log D(G(z^{(i)}; \theta_{G});  \theta_{D} ).   
	\end{equation}
\end{small}where $G(z^{(i)})$ is the generated image, $D(G(z^{(i)}))$ represents the probability of the discriminator over the prediction, and $\lambda $ is the trade-off weight, which is set by the cross-validation experiments.

\subsection{Heuristics Module} \label{heuristics} 
We suppose a sequence of $N$ tasks to be learned in order, $T=\{T_{1},...,T_{N}\}$. Each task $T_{i}$ is given a dataset of $N_{i}$ paired instances, $T_{i}=\{\mathcal{X}_{i,j},\mathcal{Y}_{i,j}\}^{N_{i}}_{j=1}$, where $\mathcal{X}_{i}$ and $\mathcal{Y}_{i}$ denote the original domain and ground truth domain respectively. In training session, the $\textit{i th}$ model $H_{i}$ learns the current task $T_{i}$ and aims to optimize the neural network to replicate $T_{i}$'s real data distribution $\mathbb{P}_{i}$. Besides, $H_{i}$ should also maintain the ability to generate competitive results on the previous tasks $T_{1},...,T_{i-1}$. However, the model has no access to the previous training data and can only use current data $\{\mathcal{X}_{i},\mathcal{Y}_{i}\}$. Unlike previous efforts mainly designed for image classifiers, image generation is typically more challenging than classification. 

As discussed in Section \ref{introduction}, the formation and update of association are mediated by potentiation and depression stimulation. If two dopamine receptors, DopR1 and DopR2, stimulate almost at the same time and repeatedly in the network, the original network will be trained into an associative network \cite{handler2019distinct}. Motivated by this mechanism, the target model can also be formulated as the interaction of potentiatively guiding the association of past knowledge and depressively allowing the learning of new episodes. We propose to preserve the initial mapping between target and source domains which captures concepts when encountering new domains. This is similar to potentiation mechanism that reconstruct connections between stimulated neurons and associative neurons.

Fig \ref{association} illustrates the heuristics module applied to our framework. First, when the basic generator is trained for the $\textit{i-1 th}$ task $T_{i-1}$, a mapping agent $M$ learns to memorize the cumulative input space and record the reconstruction mapping, $\psi_{i-1}:\mathcal{X}_{i-1}\rightarrow \mathcal{Y}_{i-1}$. Then, for the $\textit{i th}$ task, the generator obtains new training paired samples $\{\mathcal{X}_{i},\mathcal{Y}_{i}\}$. A controller $C$ randomly selects mappings from the mapping memory to perform inverse mapping which reconstructs the current input space, $\psi^{-1}_{i}:\mathcal{Y}_{i}\rightarrow \{\mathcal{X}_{j}\}_{j=1}^{i-1}$. It associates part of current ground truth with the training samples of previous tasks, which avoid saving original images or extracted features. Moreover, the data distribution $\mathbb{P}_{i}$ itself will experience gradual generalization of concepts to new domains without forgetting the past distribution $\mathbb{P}_{past}$ at any moment. In other words, the generative model is continuously capturing and updating knowledge about past domains. Therefore, the objective function of the $\textit{i th}$ task can be formulated from equation (\ref{gan}):
\begin{figure*}
	\centering
	\includegraphics[scale=0.6]{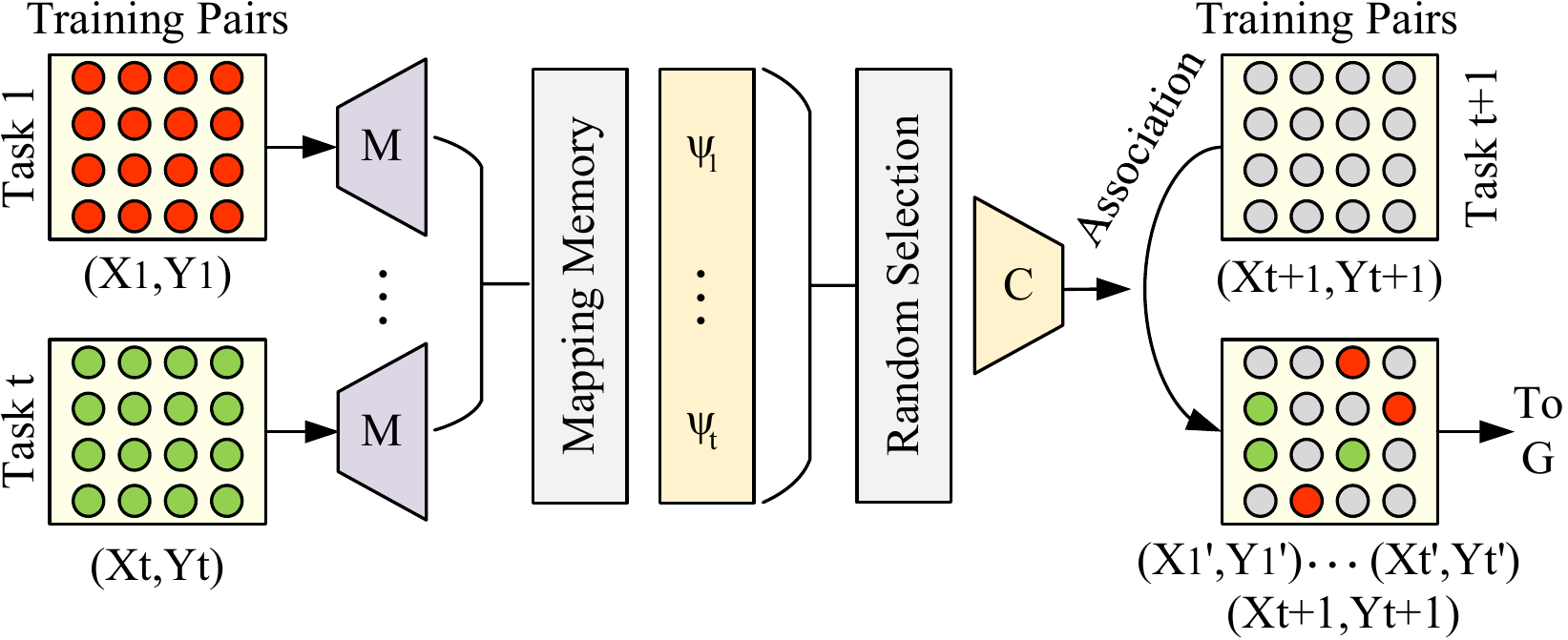}
	\caption{Heuristics module captures domain mappings from training pairs of each task and stores them into a memory structure. Then it applies the inverse mappings from the memory to reconstruct tensors and remaps them to the previous task domains. The association features will participate in the basic generator training along with the original features.}
	\label{association}
\end{figure*}

\begin{small}
	\begin{equation}
		\min \limits_{\theta_{i}} \{(1-r)\mathbb{E}_{\{\mathcal{X}_{i},\mathcal{Y}_{i}\}\sim \mathbb{P}_{i}}[\mathcal{L}(G^{i}(\mathcal{X}_{i};\theta_{i}),\mathcal{Y}_{i})]  +r\mathbb{E}_{\{\mathcal{X}^{'}_{i},\mathcal{Y}_{i}\}\sim \mathbb{P}_{past}}[\mathcal{L}(G^{i}(\psi^{-1}(\mathcal{Y}_{i};\theta_{i})),\mathcal{Y}_{i})]\}.
		\label{loss}
	\end{equation}
\end{small}where $\theta_{i}$ are parameters of the $\textit{i th}$ generator network and r is the association ratio at the current task training data. 

The proposed module can employ any deep network architecture, and we utilize the network where the architecture is the same as \cite{zhu2017unpaired} to obtain memorized mappings for association selection.

\subsection{Continual Distillation Loss}
The heuristics module is only updated along the single side trajectory from $\mathbb{P}_{i}$ to $\mathbb{P}_{past}$ during the $\textit{i th}$ task training process. Following the work on knowledge distillation for continual discriminative problem \cite{li2017learning}, we adopt a new regularization term to perform depression in the generative network. This term is similar to human depression mechanism that the plasticity of synapses is reduced to alter the efficacy of synaptic transmission. Our distillation loss aims to reduce the empirical risk between current samples and past ones, while the historical knowledge is more effectively retained by heuristics memory system as discussed in Section \ref{heuristics} . To determine the implementation of regularization and to find the vital synapses, we will rethink the process of neural network optimization from the probabilistic perspective. 

For joint learning, parameter optimization is tantamount to finding the most probable data distribution for the model given all task data $\mathcal{D}$. We can use Bayes'rule to calculate the conditional probability $p(\theta|\mathcal{D})$:

\begin{small}
	\begin{equation}
		\log p(\theta|\mathcal{X},\mathcal{Y})=\log p(\mathcal{X},\mathcal{Y}|\theta)+\log p(\theta) -\log p(\mathcal{X},\mathcal{Y}),
		\label{joint}
	\end{equation}
\end{small}where $p(\mathcal{X},\mathcal{Y})$ is the probability of data and $\log p(\theta)$ is the prior probability of parameters that can match all tasks.

For continual learning, all task data is broken up into $n$ batches $\mathcal{D}=\{\mathcal{D}_{1},...,\mathcal{D}_{n}\}$. We can derive the full objective for continual learning from equation (\ref{joint}):

\begin{small}
	\begin{equation}
		\log p(\theta_{i}|\mathcal{X},\mathcal{Y})=\log p(\mathcal{X}_{i},\mathcal{Y}_{i}|\theta_{i}) -\log p(\mathcal{X}_{i},\mathcal{Y}_{i}) +\log p(\theta_{i}|\mathcal{X}_{past},\mathcal{Y}_{past}),
	\end{equation}
\end{small}where $\{\mathcal{X}_{i},\mathcal{Y}_{i}\}$ is the data distribution of the $i th$ task, and $\{\mathcal{X}_{past},\mathcal{Y}_{past}\}$ is the data distribution of the previous i-1 tasks. Note that the probability of the $i th$ task data $p(\mathcal{X}_{i},\mathcal{Y}_{i})$ is a constant value, and the log probability of the $i th$ task data given the $i th$ task parameters $\log p(\mathcal{X}_{i},\mathcal{Y}_{i}|\theta_{i})$ is simply tantamount to the preliminary loss function at hand $\mathcal{L}$. 

$\log p(\theta_{i}|\mathcal{X}_{past},\mathcal{Y}_{past})$ is a posterior probability distribution term and it contains information about all previous task data, which is difficult to calculate. Using the theorem of diagonalized Laplace approximation \cite{mackay1992practical}, this term can be approximated as a Gaussian distribution. The mean value is the parameters of $\theta^{*}_{i-1}$ and the variance is the diagonal reciprocal of the Fisher information matrix $I$ corresponding to the parameters. Given this approximation, the optimization problem on continual distillation can be defined as the transfer of knowledge from past distribution $\mathbb{P}_{past}$ to refined distribution $\mathbb{P}_{i}$ for approximation, which is formulated as:

\begin{small}
	\begin{equation}
		\mathcal{L}_{feature}=\sum_{j}\frac{1}{2}I_{j}(\theta_{i,j}-\theta^{*}_{i-1,j})^{2},
	\end{equation}
\end{small}

Specifically, after task $T_{i-1}$ is trained, the parameters $\theta_{i-1}$ of model $G^{i-1}$ will be saved as the existing knowledge for the next task $T_{i}$. This term transfers feature knowledge between feature spaces of different tasks, alleviating over drifting in feature domains. The continual distillation loss guides the parameters of current circumstance $\theta_{i}$ to be similar to those shared by past task $\theta_{i-1}$, and this regularization term is an additional distillation for feature rather than logits. It only depends on the previous task parameters $\theta_{i-1}$ without additional storage throughout the whole training process. 

We combine the continual distillation loss with the basic loss, and the new loss function of our framework can be formulated as:

\begin{small}
	\begin{equation}
		\mathcal{L^{'}}=\mathcal{L}+\lambda^{'} \mathcal{L}_{feature}.
		\label{newloss}
	\end{equation}
\end{small} where $\lambda^{'}$ is the new trade-off weights.

The modified loss function enforces the model to update parameters to reconstruct the old experience as well as the new one. It preserves relevant past experience and generalize the concept to new domains. Therefore, the model can still continually retain the distribution of past domains $\mathbb{P}_{past}$ while integrating new tasks.

\section{Experiment}
\subsection{Comparison Models} 
We evaluate the performance of our model, Assoc-GAN, together with the representative methods, Wu et al.\cite{wu2018memory}, Lesort et al. \cite{lesort2019generative}, Lifelong VAEGAN \cite{ye2020learning}, PiggybackGAN \cite{zhai2020piggyback}, CLGAN \cite{seff2017continual}, Liang et al. \cite{liang2018generative}, LifelongGAN \cite{zhai2019lifelong}, LiSS \cite{schmidt2020towards} and Thanh et al. \cite{thanh2020catastrophic}. All these continual learning models are targeted at generative tasks, which have been introduced in Section \ref{relatedwork} . Besides, we design the following baseline methods and compare our model with them: (a) Joint Learning (JL). Data of all tasks is learned together. This is the theoretical upper bound of continual learning \cite{wu2018memory}. (b) Transfer Learning (TL). The data of each task is fed into the model sequentially for training. We use the parameters obtained from the previous task to initialize the current task model, so that preserved knowledge can be fully leveraged. This is the theoretical  lower bound of continual learning \cite{wu2018memory}. Note that \cite{liang2018generative} randomly select samples of previous tasks and store them in raw format. When the model learns new episodes, previous tasks samples will be replayed. This is the upper bound performance of memory replay method with unlimited number of examples \cite{abati2020conditional}.
\begin{figure*}
	\centering
	\subfigure[JL]{
		\includegraphics[scale=0.36]{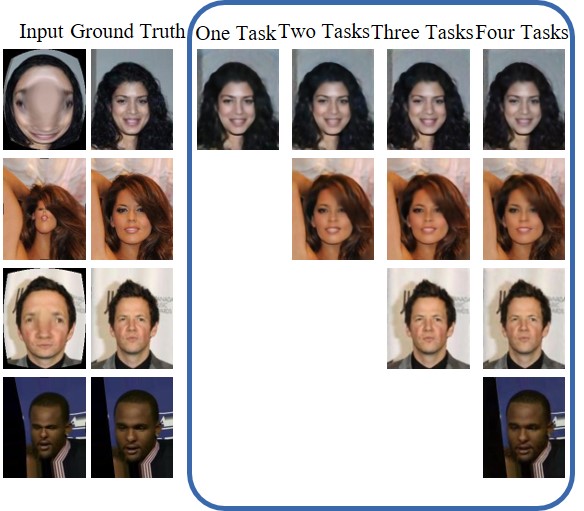}
	}\hspace{-2.5mm}
	\subfigure[TL]{
		\includegraphics[scale=0.36]{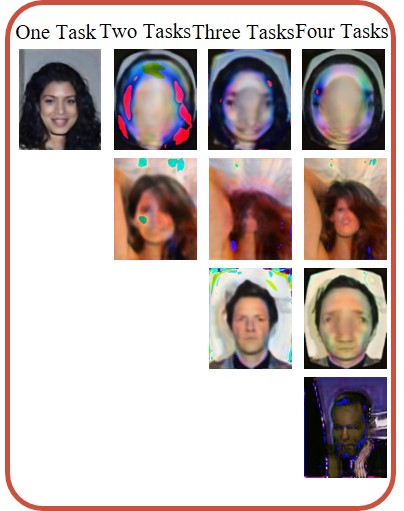}
	}\hspace{-2.5mm}
	\subfigure[\cite{wu2018memory}]{
		\includegraphics[scale=0.36]{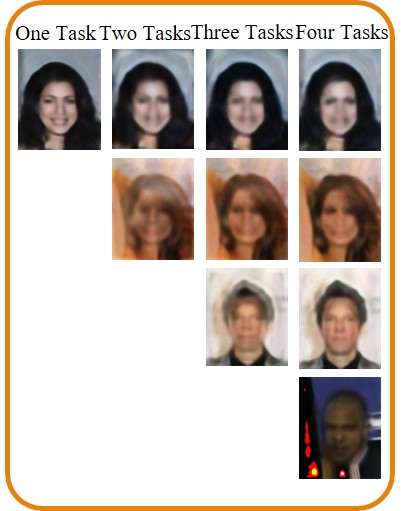}
	}\hspace{-2.5mm}
	\subfigure[\cite{lesort2019generative}]{
		\includegraphics[scale=0.36]{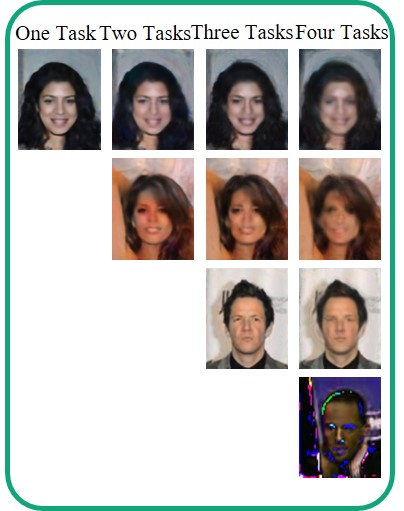}
	}\hspace{-2.5mm}
	\subfigure[LVAEGAN]{
		\includegraphics[scale=0.36]{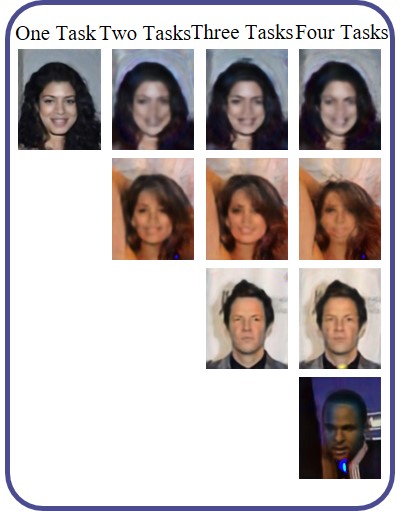}
	}\hspace{-2.5mm}
	\subfigure[PiggybackGAN]{
		\includegraphics[scale=0.36]{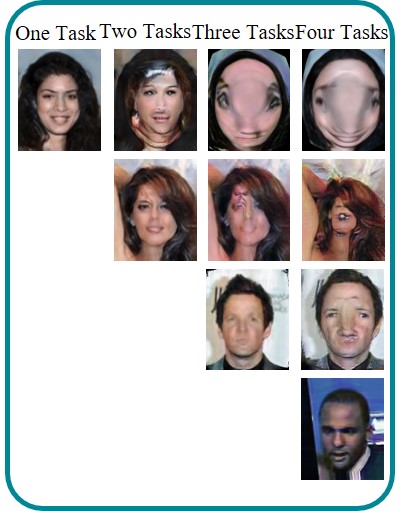}
	}\hspace{-2.5mm}
	\subfigure[LiSS]{
		\includegraphics[scale=0.36]{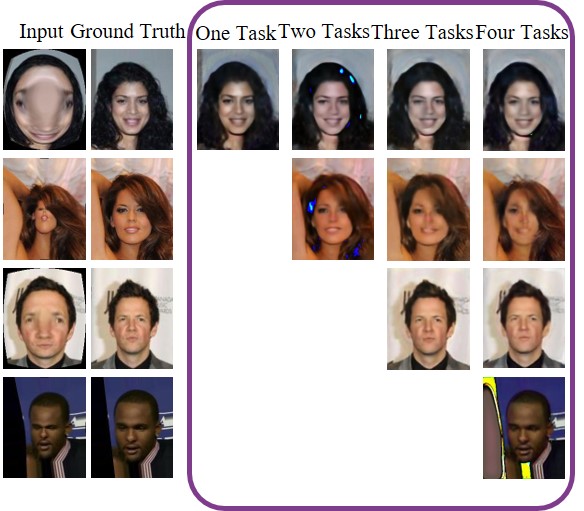}
	}\hspace{-2.5mm}
	\subfigure[CLGAN]{
		\includegraphics[scale=0.36]{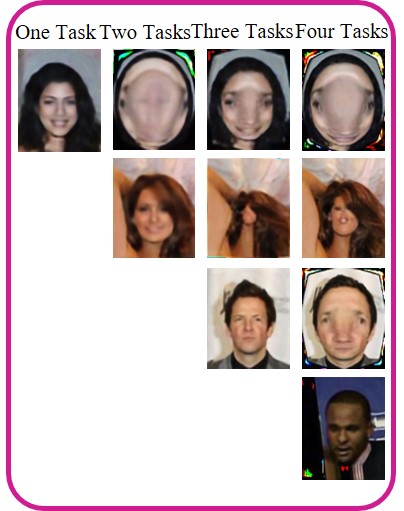}
	}\hspace{-2.5mm}
	\subfigure[\cite{liang2018generative}]{
		\includegraphics[scale=0.36]{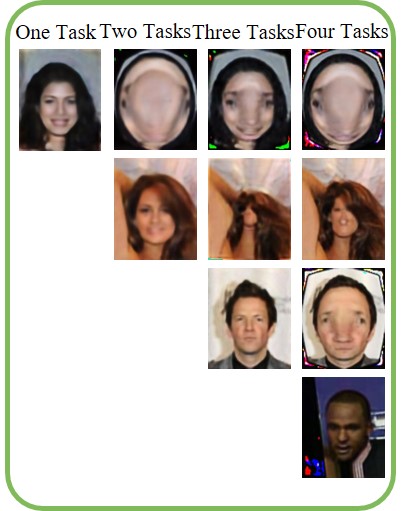}
	}\hspace{-2.5mm}
	\subfigure[LifelongGAN]{
		\includegraphics[scale=0.36]{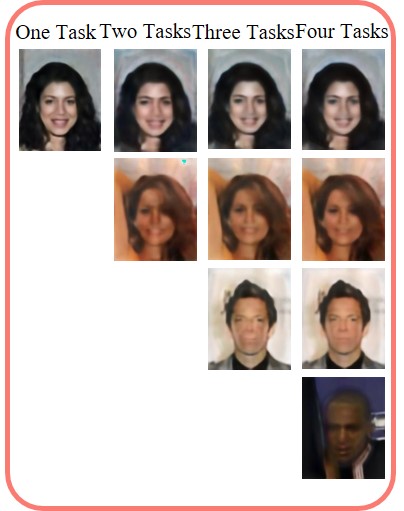}
	}\hspace{-2.5mm}
	\subfigure[\cite{thanh2020catastrophic}]{
		\includegraphics[scale=0.36]{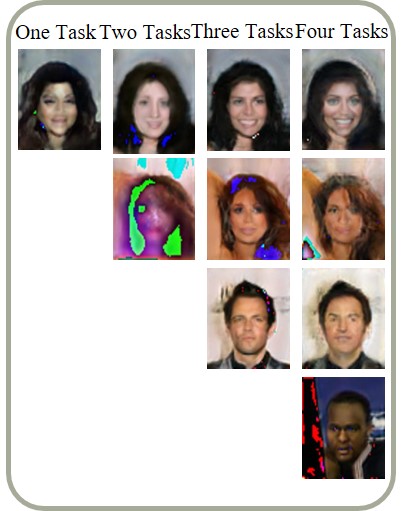}
	}\hspace{-2.5mm}
	\subfigure[Assoc-GAN]{
		\includegraphics[scale=0.36]{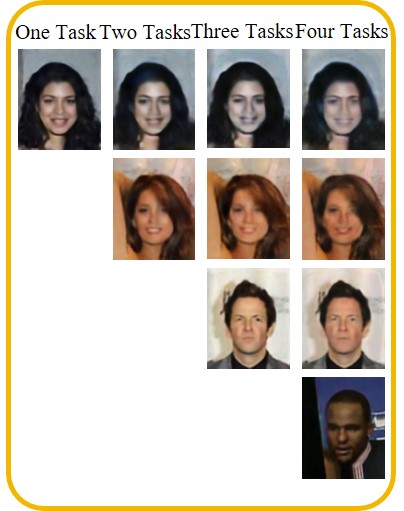}
	}
	\caption{Comparison among different methods for continual learning on DFD.}	
	\label{comparison}
\end{figure*}

\subsection{Implement Details}
We employ two benchmark datasets for our experiments: Distorted Face Dataset (DFD) \cite{gu2020generative} and Google Landmarks Dataset (GLD) \cite{weyand2020google}. DFD is a challenging reconstruction task for distorted face restoration. We split it into four tasks according to distortion types and degrees. GLD is used for image translation task. We will carry out successive style transfer on three different tone pictures. In this work, we evaluate the quality of generated prediction through qualitative and quantitative analysis. The qualitative evaluation mainly relies on visual perception while the quantitative evaluation utilizes peak signal to noise ratio (PSNR) and structural similarity (SSIM) to quantify the performance. PSNR measures the difference between corresponding pixel values and we adopt it to assess DFD. Since there are often cases in style transfer task that the evaluation result is inconsistent with the subjective feeling of the person, we adopt SSIM to assess GLD, which measures the holistic similarity from three aspects: brightness, contrast, and structure. All experiments are conducted on Tesla V100.
\begin{table}[h]
	\centering
	\scriptsize
	\resizebox{0.45\linewidth}{!}{
		\begin{tabular}[scale=0.5]{|c|c|cccc|}
			\hline
			\multicolumn{2}{|c|}{ \multirow{2}*{}} & \multicolumn{4}{|c|}{PSNR} \\ 
			\multicolumn{2}{|c|}{}& One task & Two tasks & Three tasks & Four tasks \\ 
			\hline
			\multirow{5}{*}{TL} & T1 & 27.02  & 9.72  & 11.97  & 10.03  \\ 
			& T2 &  & 20.64  & 12.78  & 13.44  \\
			& T3 &  &  & 20.06  & 11.99  \\
			& T4 &  &  &  & 20.67  \\
			& AVG & 27.02  & 15.18  & 14.94  & 14.03  \\ \hline
			\multirow{5}{*}{JL} & T1 & 26.15  & 27.85  & 28.36  & 26.59  \\
			& T2 &  & 26.92  & 27.22  & 26.52  \\ 
			& T3 &  &  & 31.23  & 28.51  \\
			& T4 &  &  &  & 30.31  \\
			& AVG & 26.15  & 27.39  & 28.94  & 27.98  \\ \hline
			\multirow{5}{*}{CLGAN} & T1 & 25.26  & 9.03  & 11.96  & 9.55  \\ 
			& T2 &  & 24.32  & 12.85  & 15.09  \\ 
			& T3 &  &  & 28.08  & 11.44  \\ 
			& T4 &  &  &  & 29.85  \\
			& AVG & 25.26  & 16.68  & 17.63  & 16.48  \\ \hline
			\multirow{5}{*}{\cite{liang2018generative}} & T1 & 25.21  & 8.99  & 11.95  & 9.58  \\
			& T2 &  & 24.17  & 12.73  & 14.99  \\ 
			& T3 &  &  & 28.08  & 11.49  \\ 
			& T4 &  &  &  & 28.88  \\
			& AVG & 25.21  & 16.58  & 17.59  & 16.24  \\ \hline
			\multirow{5}{*}{\cite{wu2018memory}} & T1 & 24.82  & 21.66  & 21.20  & 20.92  \\ 
			& T2 &  & 17.48  & 20.06  & 21.53  \\ 
			& T3 &  &  & 20.08  & 22.81  \\
			& T4 &  &  &  & 21.55  \\
			& AVG & 24.82  & 19.57  & 20.45  & 21.70  \\ \hline
			\multirow{5}{*}{LifelongGAN} & T1 & 26.50  & 21.96  & 22.36  & 21.84  \\
			& T2 &  & 20.58  & 21.75  & 21.92  \\
			& T3 &  &  & 22.53  & 23.38  \\ 
			& T4 &  &  &  & 22.68  \\
			& AVG & 26.50  & 21.27  & 22.21  & 22.45  \\ \hline
			\multirow{5}{*}{\cite{lesort2019generative}} & T1 & 26.20  & 22.74  & 23.54  & 23.12  \\ 
			& T2 &  & 22.74  & 23.31  & 20.19  \\ 
			& T3 &  &  & 23.84   & 25.23  \\
			& T4 &  &  &  & 23.44  \\
			& AVG & 26.20  & 22.74  & 24.49  & 23.00  \\ \hline
			\multirow{5}{*}{LiSS} & T1 & 23.26  & 21.47  & 19.61  & 21.06  \\ 
			& T2 &  & 22.71  & 21.95  & 17.99  \\ 
			& T3 &  &  & 28.08  & 24.69  \\
			& T4 &  &  &  & 27.52  \\
			& AVG & 23.26  & 22.09  & 23.21  & 22.82  \\ \hline
			\multirow{5}{*}{LVAEGAN} & T1 & 23.32  & 21.70  & 20.15  & 19.90  \\
			& T2 &  & 23.27  & 22.13  & 18.53  \\ 
			& T3 &  &  & 26.57  & 24.05  \\
			& T4 &  &  &  & 26.82  \\ 
			& AVG & 23.32  & 22.49  & 22.95  & 22.33  \\ \hline
			\multirow{5}{*}{PiggybackGAN} & T1 & 25.75  & 15.07  & 12.86  & 12.84  \\
			& T2 &  & 30.74  & 14.37  & 13.23  \\ 
			& T3 &  &  & 29.27  & 14.72  \\ 
			& T4 &  &  &  & 26.37  \\
			& AVG & 25.75  & 22.91  & 18.83  & 16.79  \\ \hline
			\multirow{5}{*}{\cite{thanh2020catastrophic}} & T1 & 25.98  & 16.65  & 19.39  & 18.38  \\
			& T2 &  & 17.78  & 19.54  & 18.20  \\ 
			& T3 &  &  & 22.51  & 22.39  \\ 
			& T4 &  &  &  & 24.85  \\
			& AVG & 25.98  & 17.22  & 20.48  & 20.95  \\ 
			\hline
			\multirow{5}{*}{Assoc-GAN} & T1 & 26.72  & 22.44  & 21.17  & 21.63  \\ 
			& T2 &  & 25.11  & 23.19  & 22.16  \\ 
			& T3 &  &  & 28.06  & 25.03  \\ 
			& T4 &  &  &  & 27.75  \\ 
			& AVG & 26.72  & 23.78  & 24.14  & 24.14  \\ \hline
		\end{tabular}
	}
	
	\caption{Quantitative evaluations on DFD in terms of PSNR. Higher values are better.}
	\label{DFD}
\end{table}

\subsection{Results: DFD}
\textbf{Perceptual Study.} Table \ref{DFD} presents the PSNR results for quantitative evaluation. It is noteworthy that, Assoc-GAN achieves PSNR of 24.14 dB on four tasks, which takes the achievement of the highest numerical performance and surpasses most continual learning models. JL is not subject to continual learning conditions, and it sets up the upper bound for all methods. TL only generates incoherent facade-like patterns and obviously completely forgets the acquired knowledge. \cite{liang2018generative} is competitive with Assoc-GAN, but it directly revisits past training data. By contrast, our model has no access to original images and still obtains promising results. Due to the difference in distortion types between tasks, the distribution of each task is different, CLGAN, \cite{liang2018generative} and Piggyback-GAN relying only on knowledge transfer have limited influence on alleviating catastrophic forgetting. Some visual examples are also shown in Figure \ref{comparison}. These results illustrate that our inference model obtains predictions indistinguishable from ground truth and works well on split DFD. Note that as the distortion degree increases, the metric decreases and more details are lost in the produced results, so T1 and T2 themselves are more intractable to restore than T3 and T4. Overall, the hallucinated results produced by Assoc-GAN are perceptually convincing and it performs particularly well on all four tasks without forgetting historical knowledge.

\begin{figure}
	\centering
	\hspace{-5mm}
	\subfigure[Memory]{
		\includegraphics[width=0.4\linewidth]{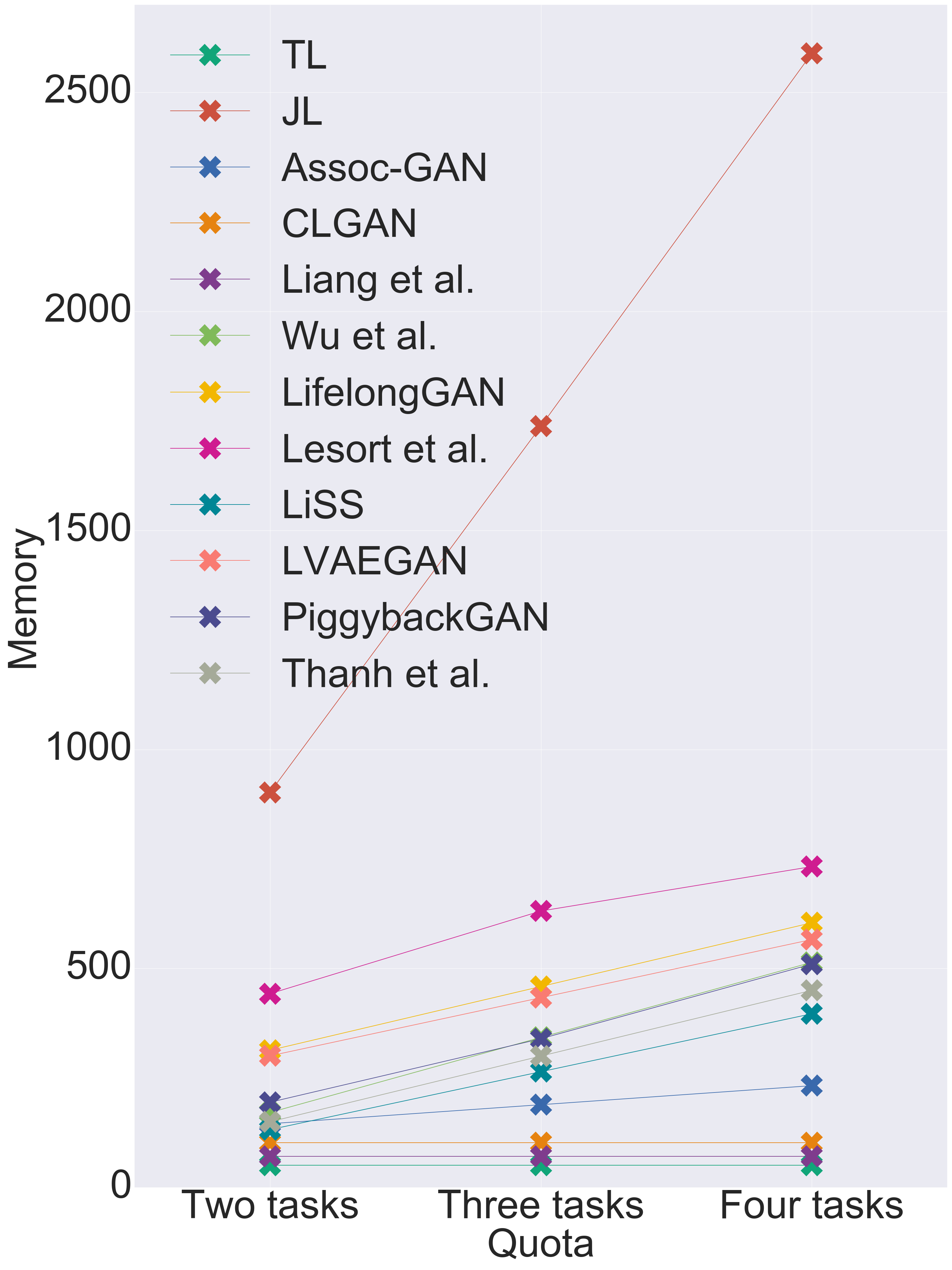}
		\label{memory}
	}\hspace{-0mm}
	\subfigure[Time]{
		\includegraphics[width=0.4\linewidth]{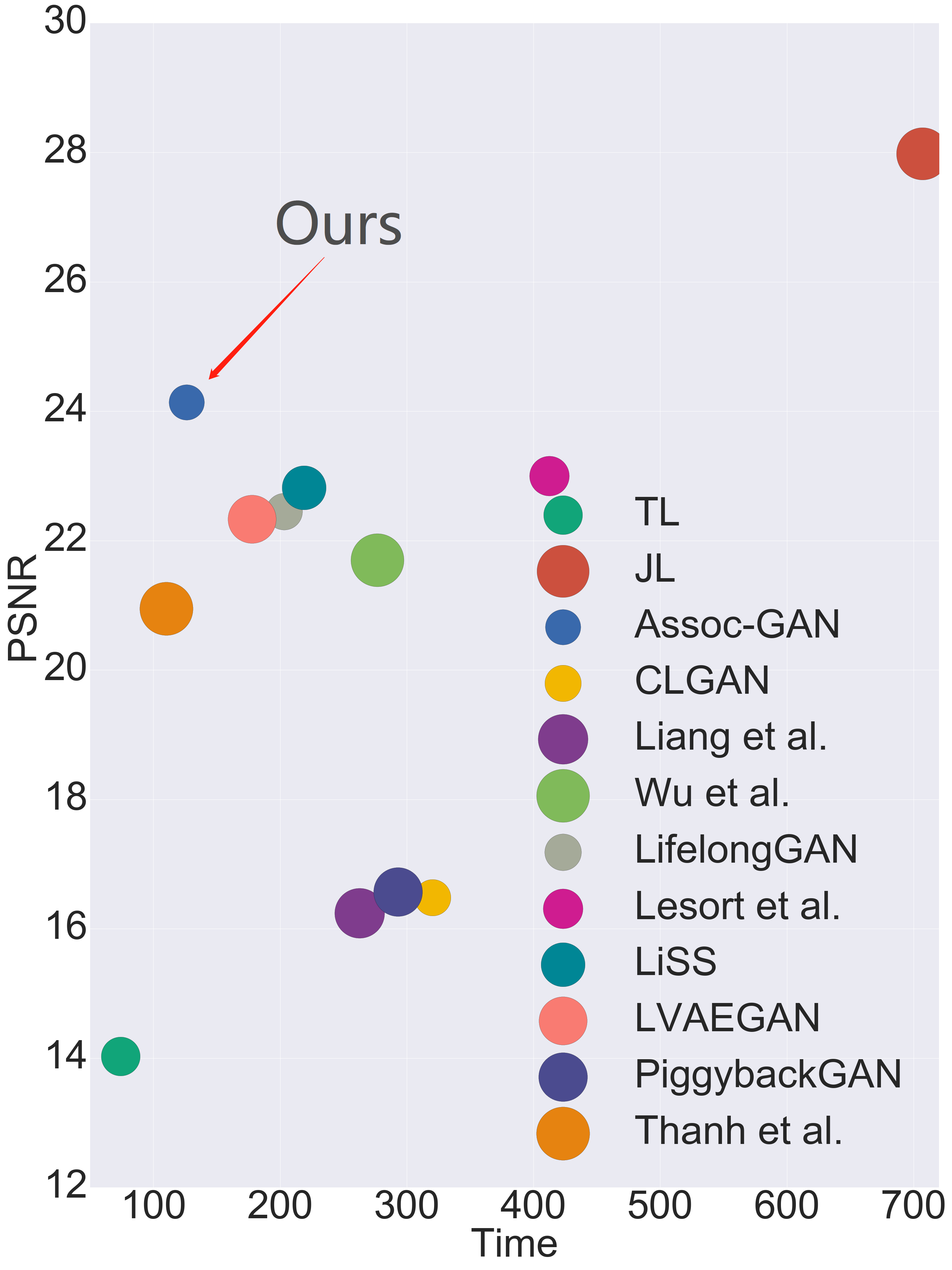}
		\label{time}
	}\vspace{-1mm}
	\caption{Time and memory study on DFD.}	
\end{figure}

\textbf{Analysis 1: Memory.} As shown in Figure \ref{memory}, we estimate the viability of memory usage. In addition to the current task data, JL, \cite{wu2018memory}, LifelongGAN, \cite{lesort2019generative}, LVAEGAN and \cite{thanh2020catastrophic} need extra memory to store the historical feature information. It can be observed that as task number increase, the memory overheads of methods above show a rapid increase, while the other methods almost come to a standstill. Our Assoc-GAN decreases the memory usage at all criteria. Specifically, when each model reaches an almost all equal state on four tasks, our memory overhead is 9.0\% of JL, 31.7\% of \cite{lesort2019generative} and 38.5\% of LifelongGAN, respectively. It is only next to CLGAN and \cite{liang2018generative} but the continual learning performance of the latter two is far less than the former. Consistently we can claim that Assoc-GAN is suitable for deployment and occupied storage size does not increase significantly. 

\textbf{Analysis 2: Time.} Training time of each method is summarized in Figure \ref{time}. The task size of each method remains consistent during training process. Assoc-GAN benefits from the advantage of heuristics module, which takes time to heuristically backtrack episodes. This process maintains the total number of samples, but only changes occupancy ratio of sample distributions for different tasks, which makes our training time irrelevant to the task size. Memory replay methods require retaining part of previous information, and their training time is related to the scale of replay samples. Only \cite{thanh2020catastrophic} trains slightly faster than Assoc-GAN, but the continual learning performance of our model far exceeds the former. We can conclude that Assoc-GAN is not cumbersome for training and the computing power overhead does not increase significantly. 

\begin{figure}
	\centering
	\subfigure[TL]{
		\includegraphics[scale=0.11]{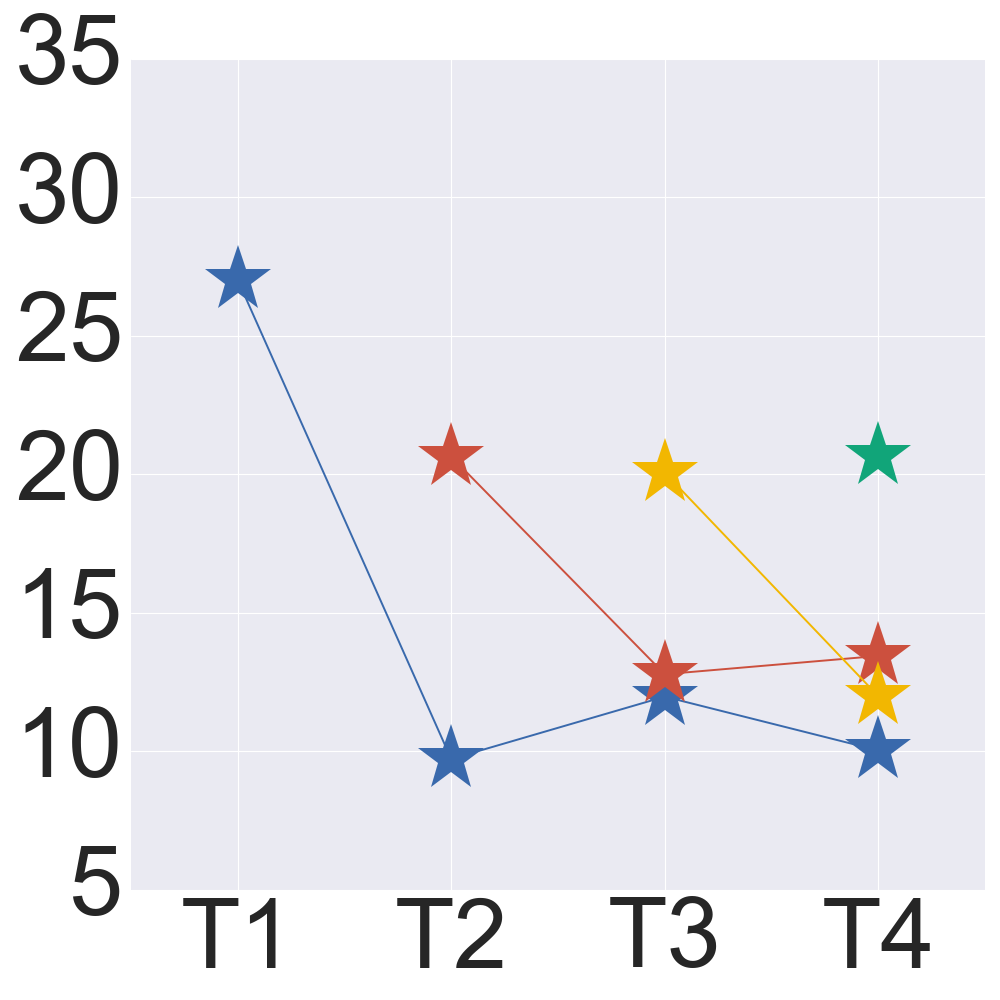}
	}\hspace{-4mm} \vspace{-4.0 mm}
	\subfigure[JL]{
		\includegraphics[scale=0.11]{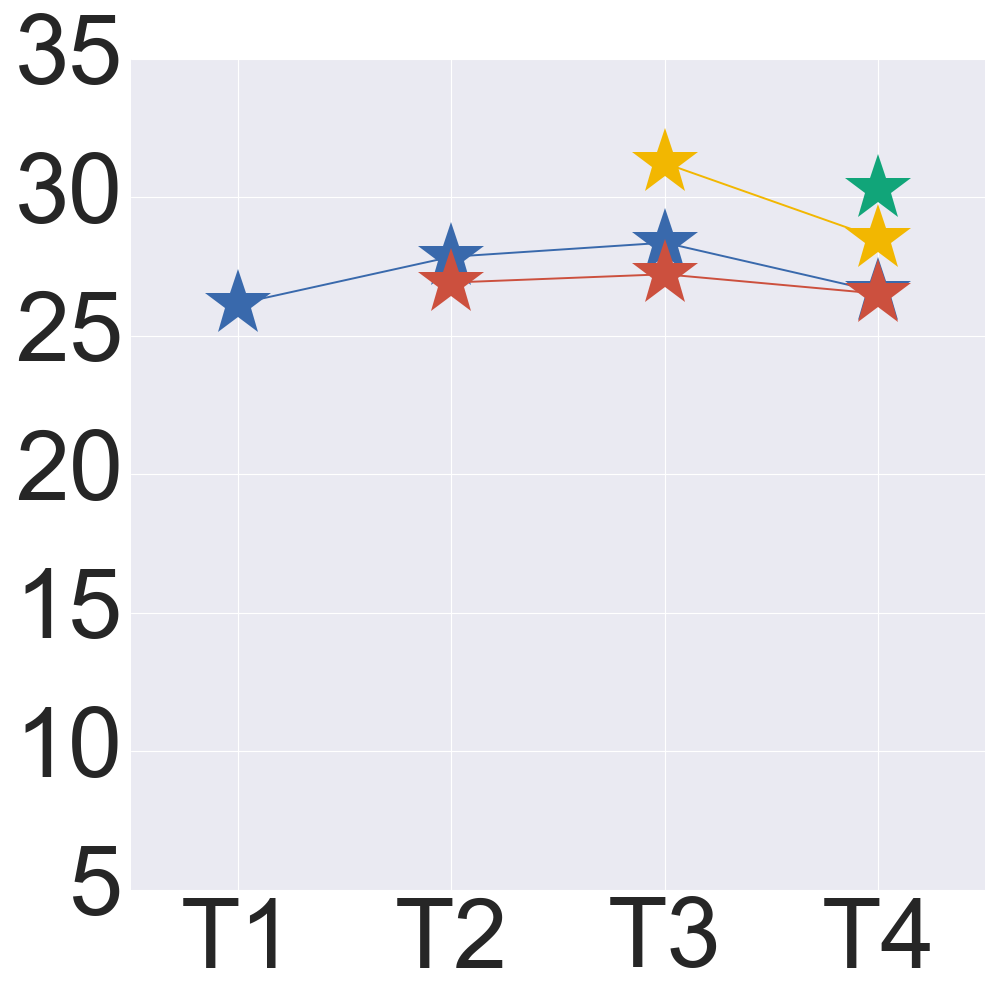}
	}\hspace{-4mm}
	\subfigure[Wu et al.]{
		\includegraphics[scale=0.11]{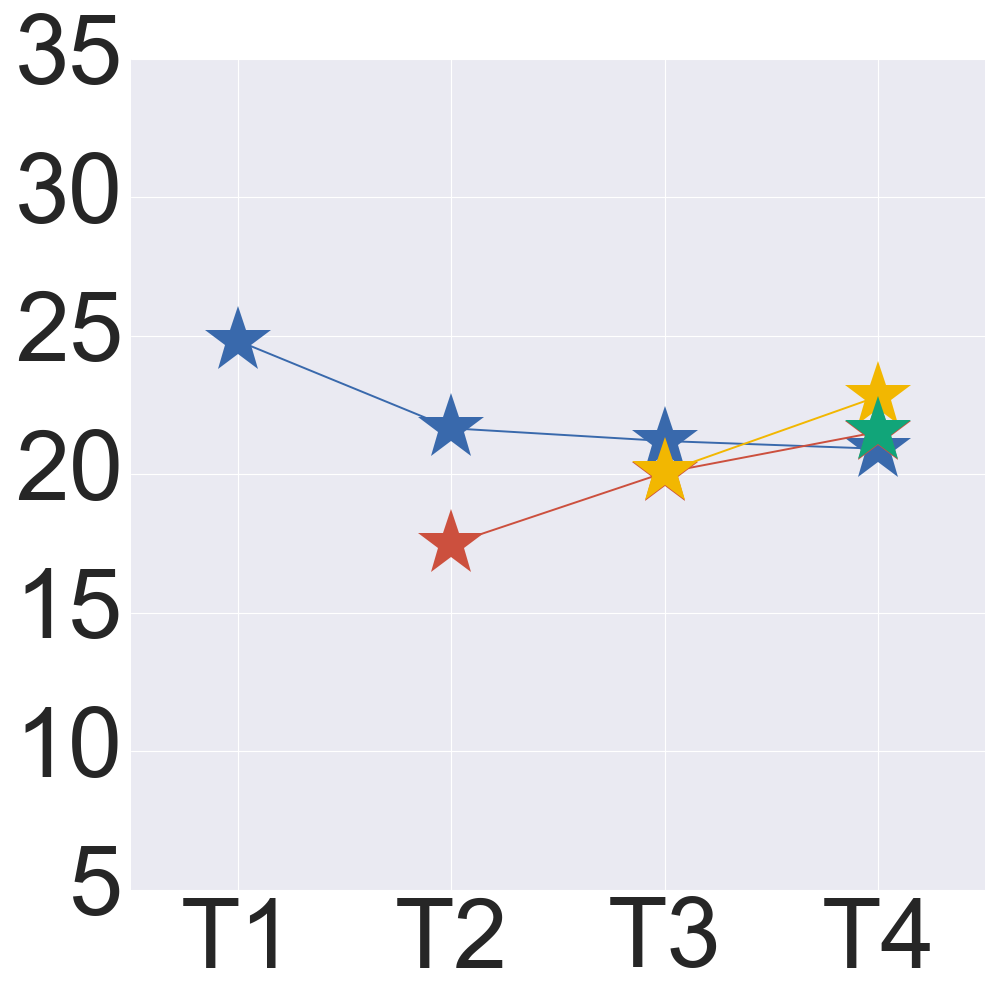}
	}\hspace{-4mm}
	\subfigure[Lesort et al.]{
		\includegraphics[scale=0.11]{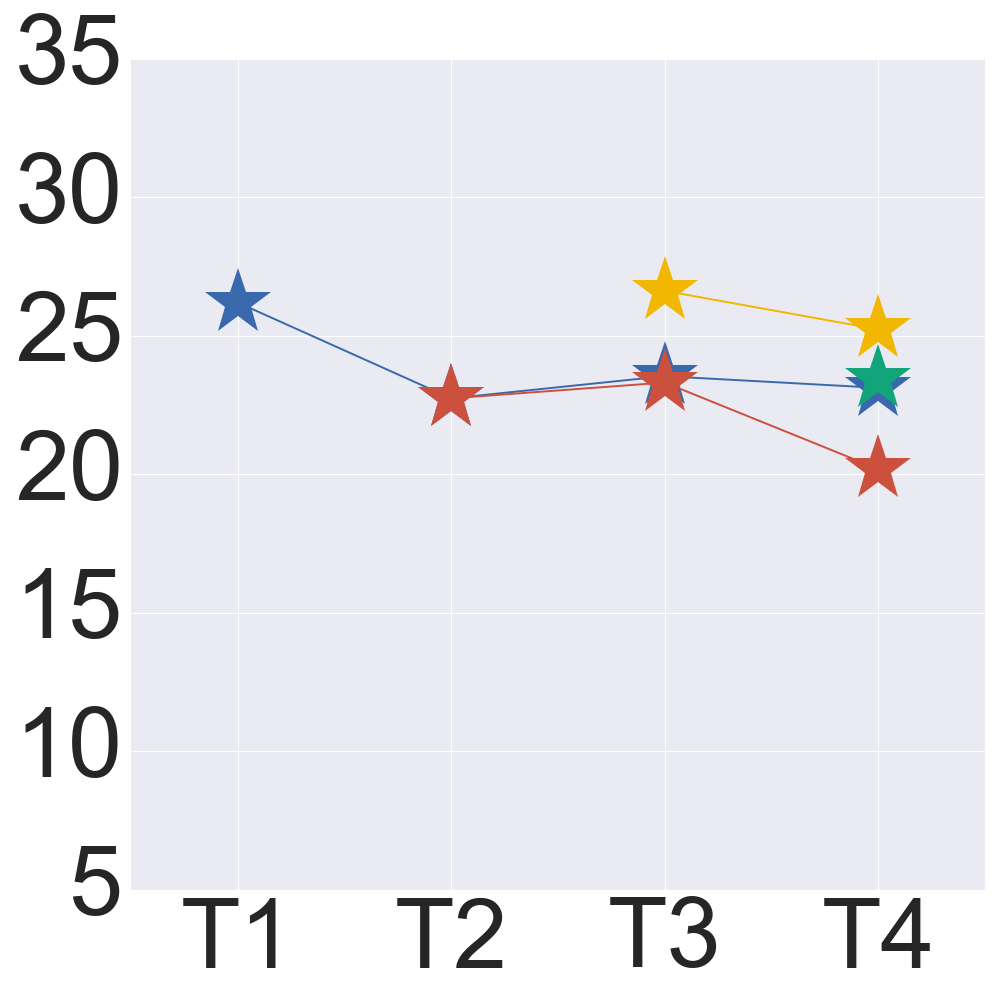}
	}\hspace{-4mm} 
	\subfigure[LVAEGAN]{
		\includegraphics[scale=0.11]{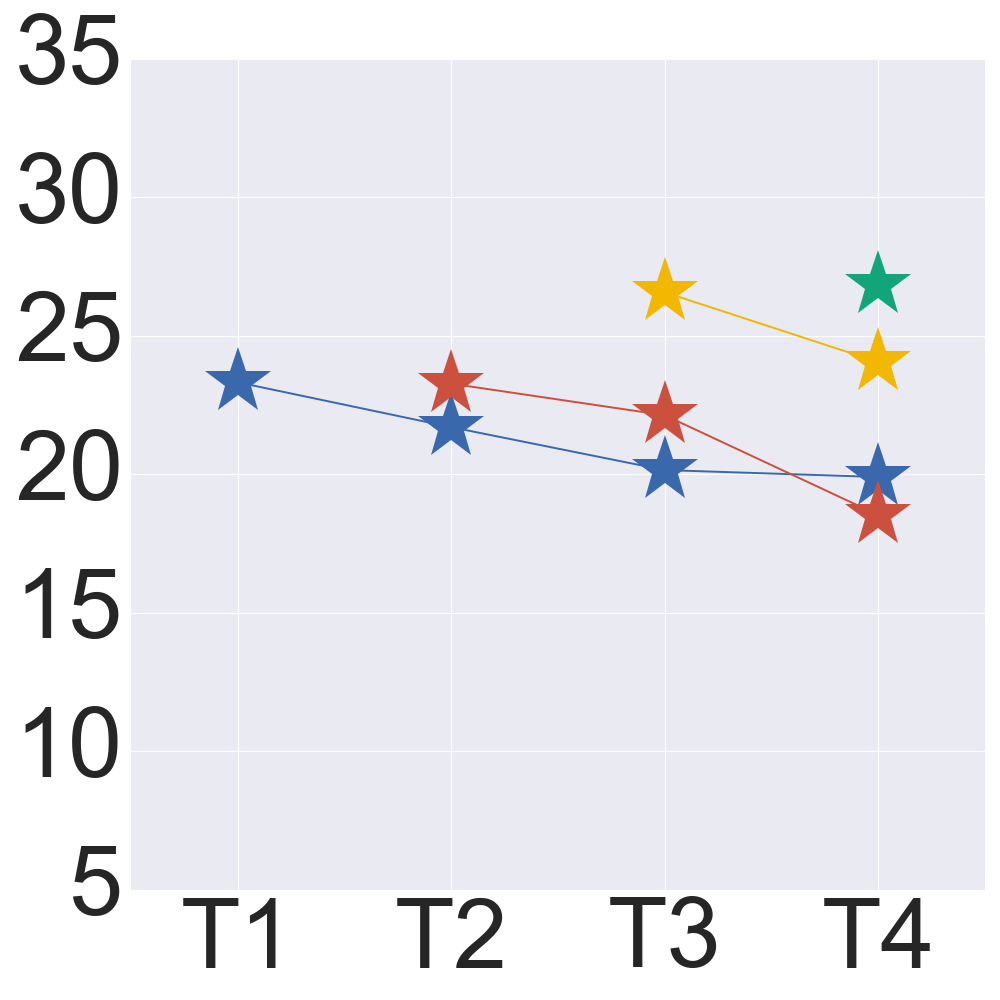}
	}\hspace{-4mm}
	\subfigure[LiSS]{
		\includegraphics[scale=0.11]{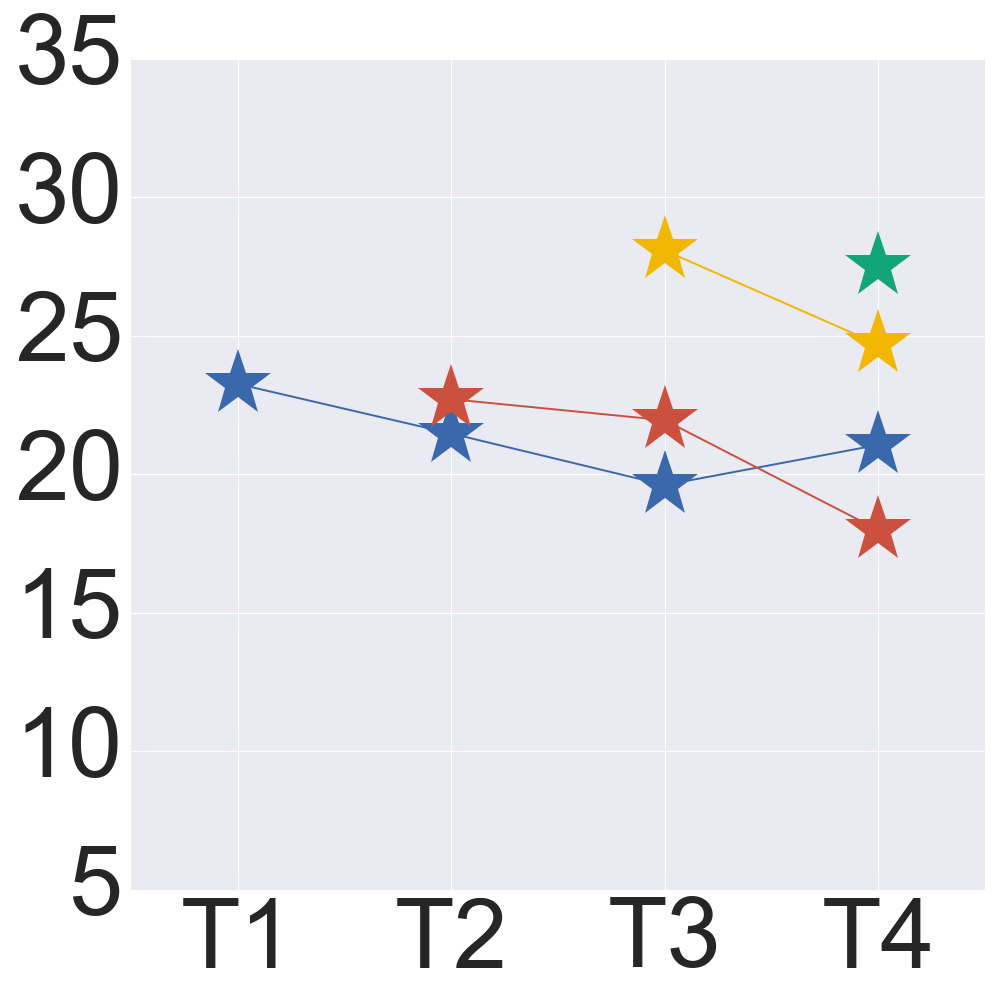}
	}\hspace{-4mm}
	\subfigure[CLGAN]{
		\includegraphics[scale=0.11]{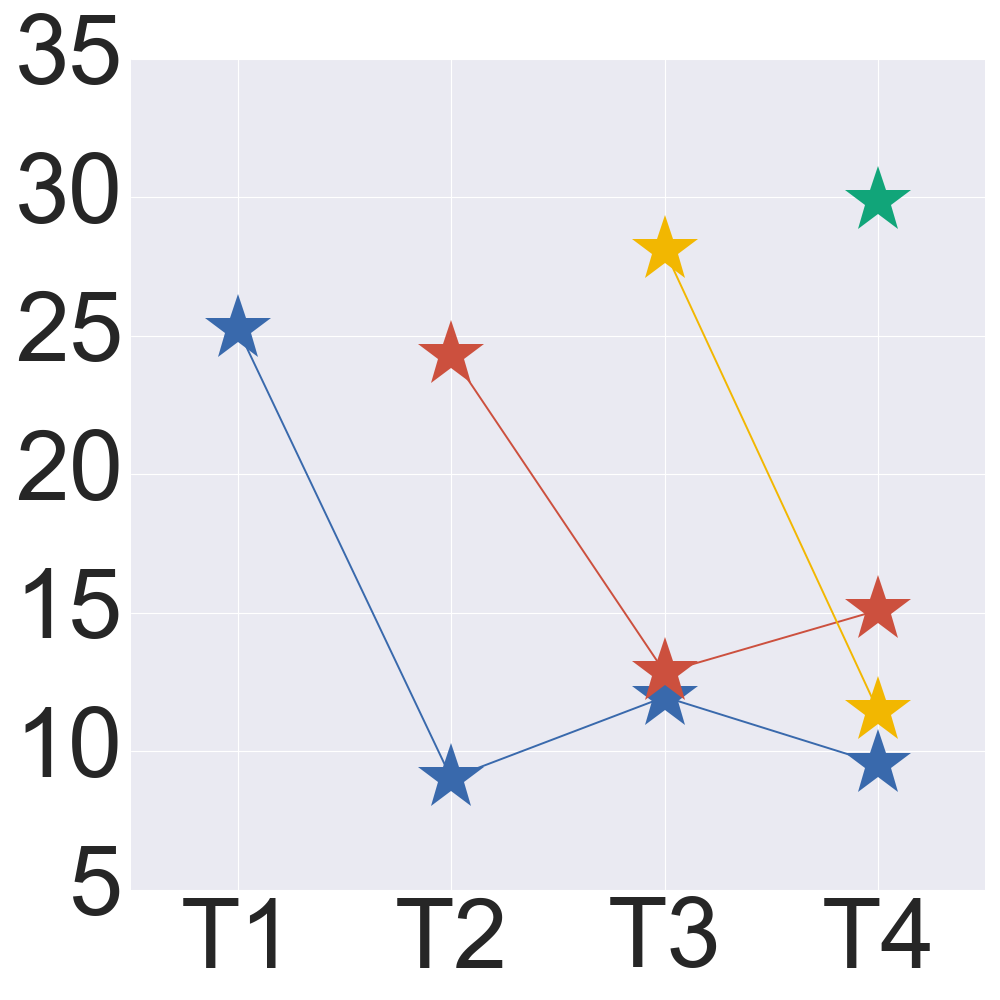}
	}\hspace{-4mm}
	\subfigure[Piggyback]{
		\includegraphics[scale=0.11]{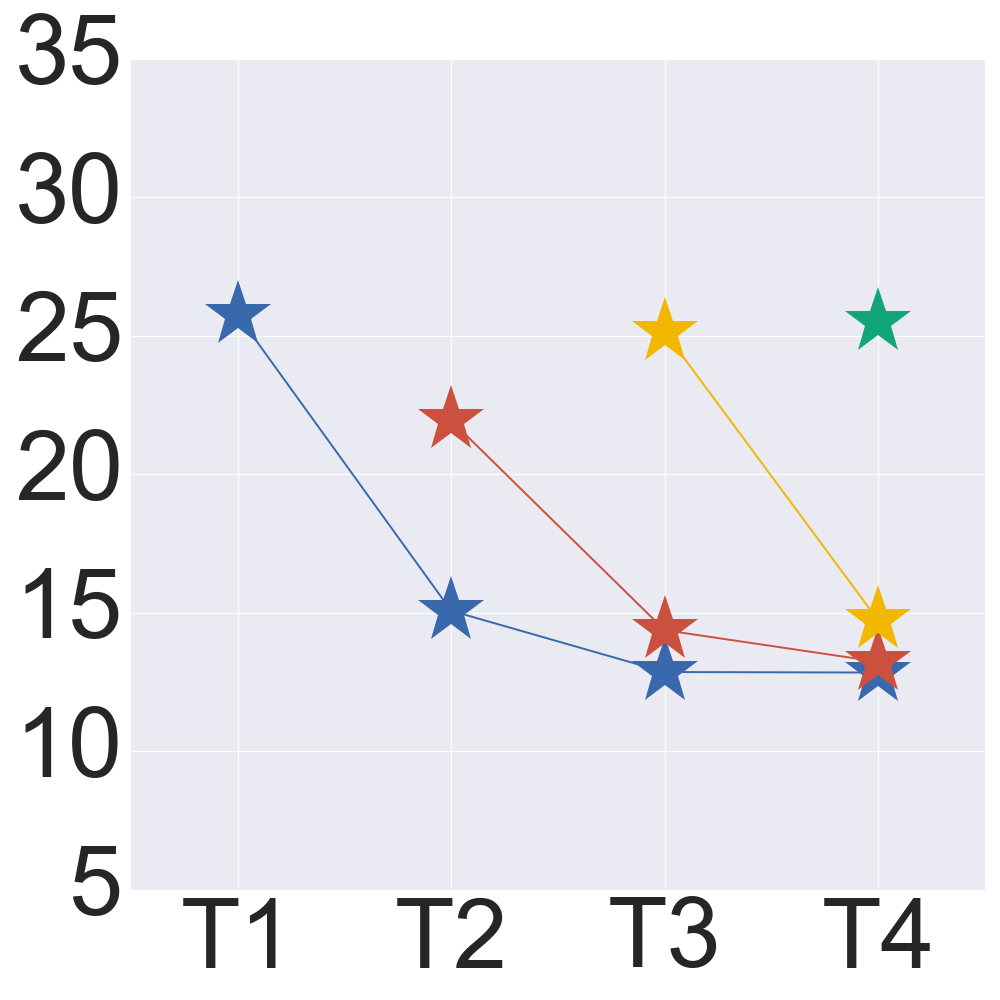}
	}\hspace{-4mm}
	\subfigure[Liang et al.]{
		\includegraphics[scale=0.11]{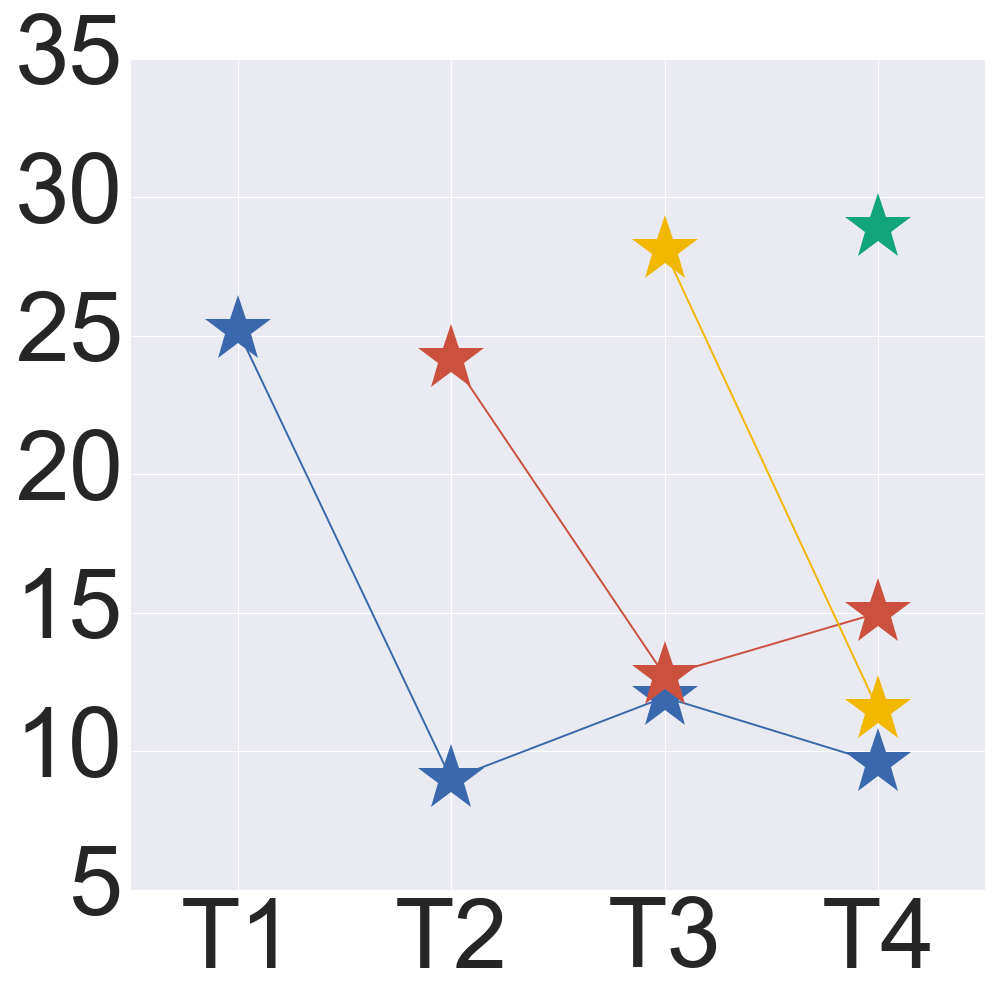}
	}\hspace{-4mm}
	\subfigure[LifelongGAN]{
		\includegraphics[scale=0.11]{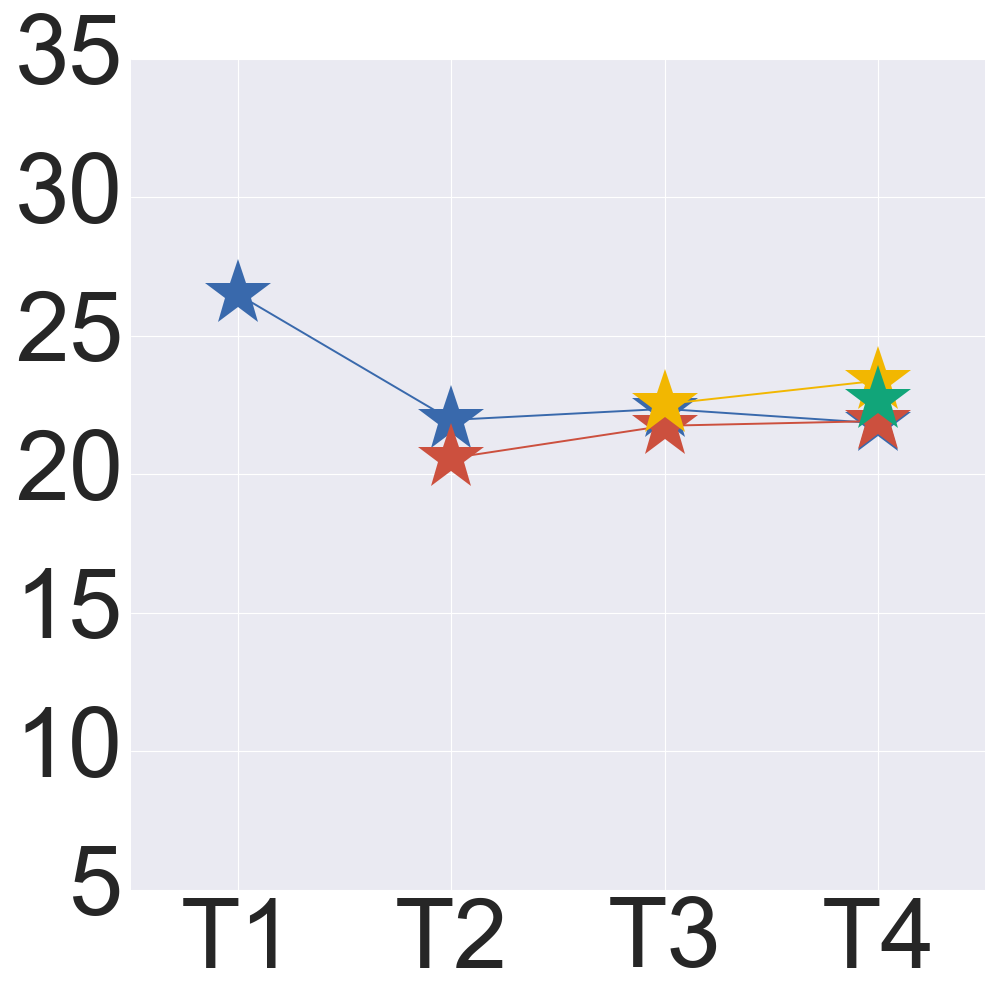}
	}\hspace{-4mm}
	\subfigure[Thanh et al.]{
		\includegraphics[scale=0.11]{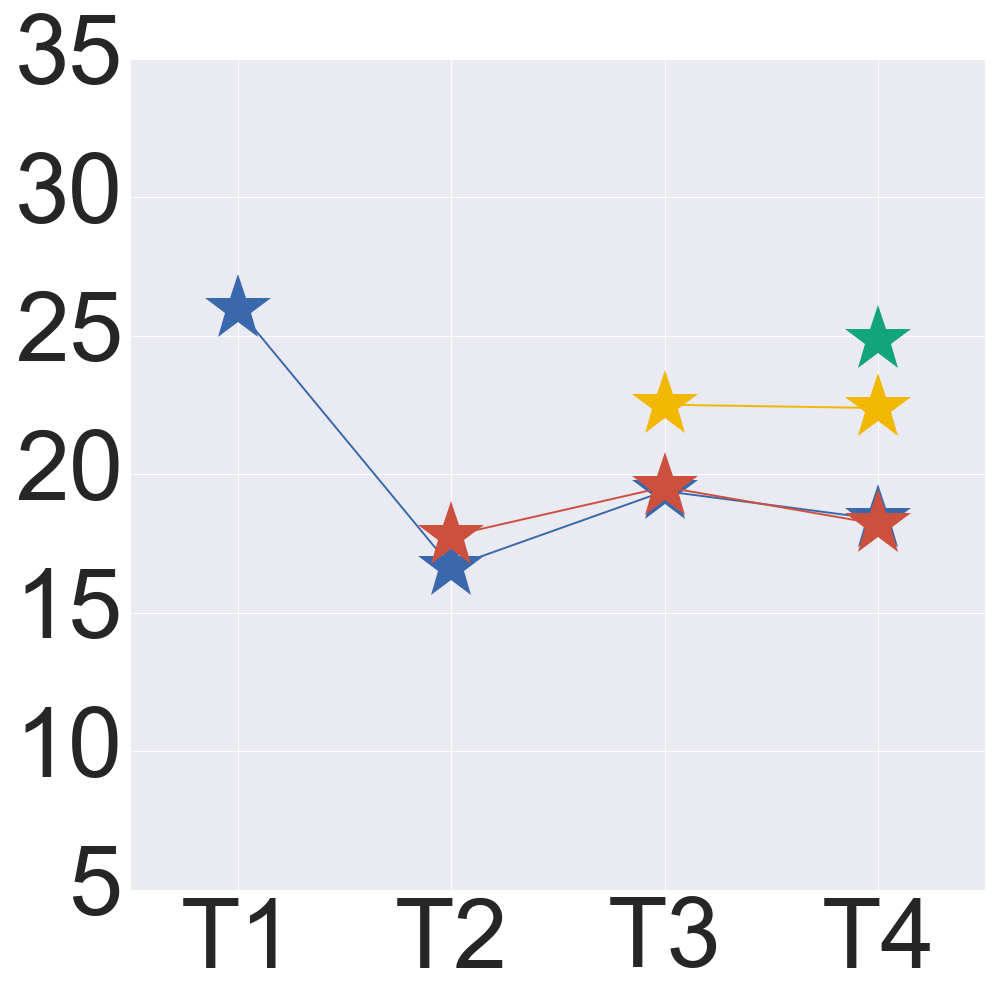}
	}\hspace{-4mm}
	\subfigure[Assoc-GAN]{
		\includegraphics[scale=0.11]{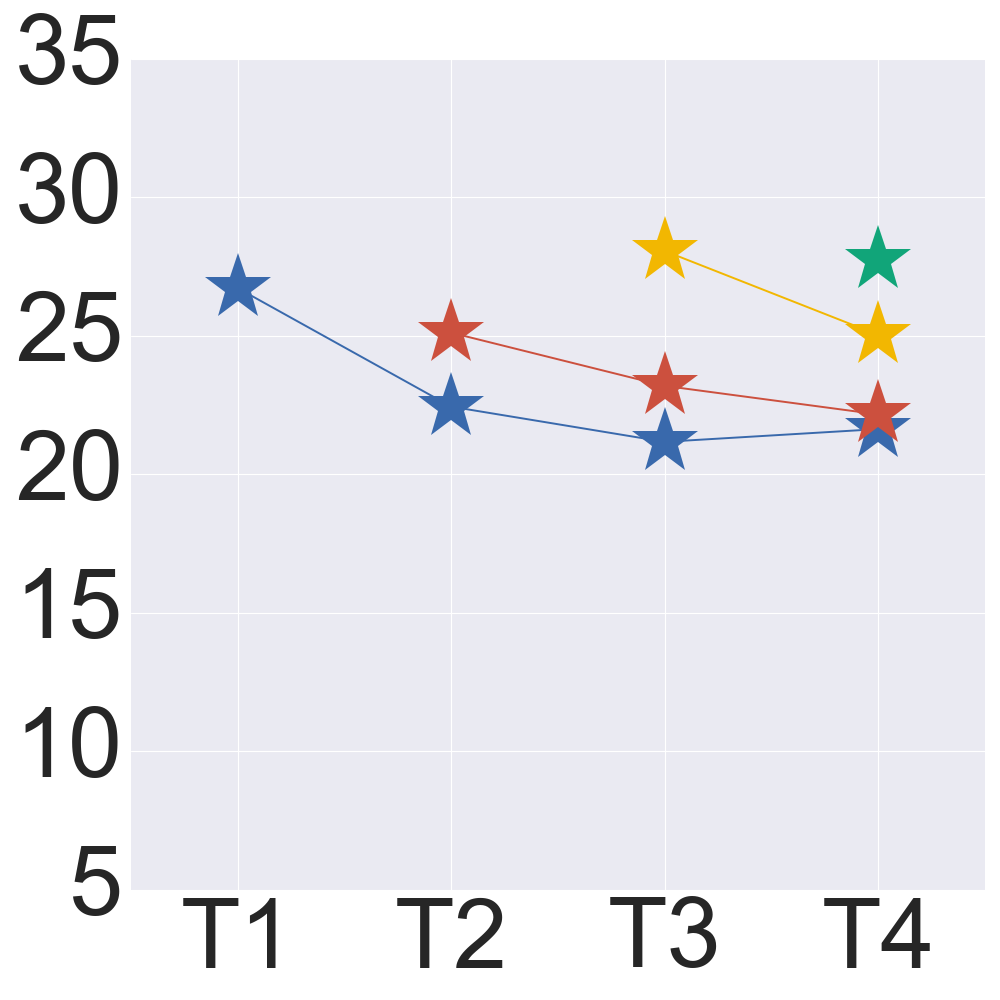}
	}
	\caption{Forgetting Analysis. The blue line is Task1, the red line is Task2, the yellow line is Task3, and the green dot is Task4.}	
	\label{forget}
\end{figure}

\textbf{Analysis 3: Forgetting.} We observe the evolution of PSNR on four tasks during the whole training process from Figure \ref{forget}. TL, CLGAN, \cite{liang2018generative}, PiggybackGAN suffer from catastrophic forgetting. Once the task is separately input, the model will inevitably forget the historical knowledge to varying degrees. By contrast, our model retain the previous task knowledge throughout sequential training on all cases and keep the performance on a high plane. It is clear that the knowledge forgetting speed under Assoc-GAN is slower than other replay methods and it draws a significantly better PSNR-quota curve compared with other models. More detailed analysis can be find in supplementary materials.

\begin{table}
	\centering
	\scriptsize
	\resizebox{0.45\linewidth}{!}{
		\begin{tabular}{|c|c|ccc|}
			\hline
			\multicolumn{2}{|c|}{}  & \multicolumn{3}{|c|}{SSIM}  \\ 
			\multicolumn{2}{|c|}{} & One task & Two tasks & Three tasks\\ \hline
			\multirow{4}{*}{TL} & T1 & 0.81  & 0.46  & 0.59  \\
			& T2 &  & 0.88  & 0.41  \\
			& T3 &  &  & 0.82  \\
			& AVG & 0.81  & 0.67  & 0.61  \\ \hline
			\multirow{4}{*}{JL} & T1 & 0.85  & 0.85  & 0.74  \\
			& T2 &  & 0.87  & 0.87  \\
			& T3 &  &  & 0.83  \\
			& AVG & 0.85  & 0.86  & 0.81  \\ \hline
			\multirow{4}{*}{CLGAN} & T1 & 0.79  & 0.46  & 0.60  \\
			& T2 &  & 0.89  & 0.45  \\
			& T3 &  &  & 0.83  \\
			& AVG & 0.79  & 0.68  & 0.63  \\ \hline
			\multirow{4}{*}{\cite{liang2018generative}} & T1 & 0.82  & 0.44  & 0.57  \\
			& T2 &  & 0.88  & 0.48  \\ 
			& T3 &  &  & 0.82  \\ 
			& AVG & 0.82  & 0.66  & 0.63  \\ \hline
			\multirow{4}{*}{\cite{wu2018memory}} & T1 & 0.80  & 0.64  & 0.71  \\
			& T2 &  & 0.84  & 0.61  \\
			& T3 &  &  & 0.77  \\
			& AVG & 0.80  & 0.74  & 0.70  \\ \hline
			\multirow{4}{*}{LifelongGAN} & T1 & 0.83  & 0.76  & 0.74  \\
			& T2 &  & 0.63  & 0.56  \\
			& T3 &  &  & 0.73  \\
			& AVG & 0.83  & 0.70  & 0.68  \\  \hline
			\multirow{4}{*}{\cite{lesort2019generative}} & T1 & 0.84  & 0.79  & 0.74  \\
			& T2 &  & 0.84  & 0.83  \\
			& T3 &  &  & 0.84  \\
			& AVG & 0.84  & 0.82  & 0.80  \\ \hline
			\multirow{4}{*}{LiSS}& T1 & 0.79  & 0.71  & 0.67  \\
			& T2 &  & 0.85  & 0.62  \\
			& T3 &  &  & 0.81  \\
			& AVG & 0.79  & 0.78  & 0.70  \\ \hline
			\multirow{4}{*}{LVAEGAN} & T1 & 0.80  & 0.71  & 0.67  \\
			& T2 &  & 0.84  & 0.57  \\
			& T3 &  &  & 0.78  \\
			& AVG & 0.80  & 0.77  & 0.67  \\  \hline
			\multirow{4}{*}{\cite{thanh2020catastrophic}} & T1 & 0.69  & 0.71  & 0.59  \\
			& T2 &  & 0.77  & 0.63  \\
			& T3 &  &  & 0.71  \\
			& AVG & 0.72  & 0.74  & 0.64  \\ \hline
			\multirow{4}{*}{PiggybackGAN} & T1 & 0.76  & 0.35  & 0.56  \\
			& T2 &  & 0.49  & 0.51  \\
			& T3 &  &  & 0.80  \\
			& AVG & 0.76  & 0.42  & 0.62  \\ \hline
			\multirow{4}{*}{Assoc-GAN} & T1 & 0.81  & 0.75  & 0.71  \\
			& T2 &  & 0.86  & 0.68  \\
			& T3 &  &  & 0.81  \\
			& AVG & 0.81  & 0.80  & 0.73  \\ \hline
	\end{tabular}}
	\caption{Quantitative evaluations on GLD in terms of SSIM. Higher values are better.}
	\label{GLD}
\end{table}

\subsection{Results: GLD}
\begin{figure}
	\centering
	\subfigure{
		\includegraphics[scale=0.3]{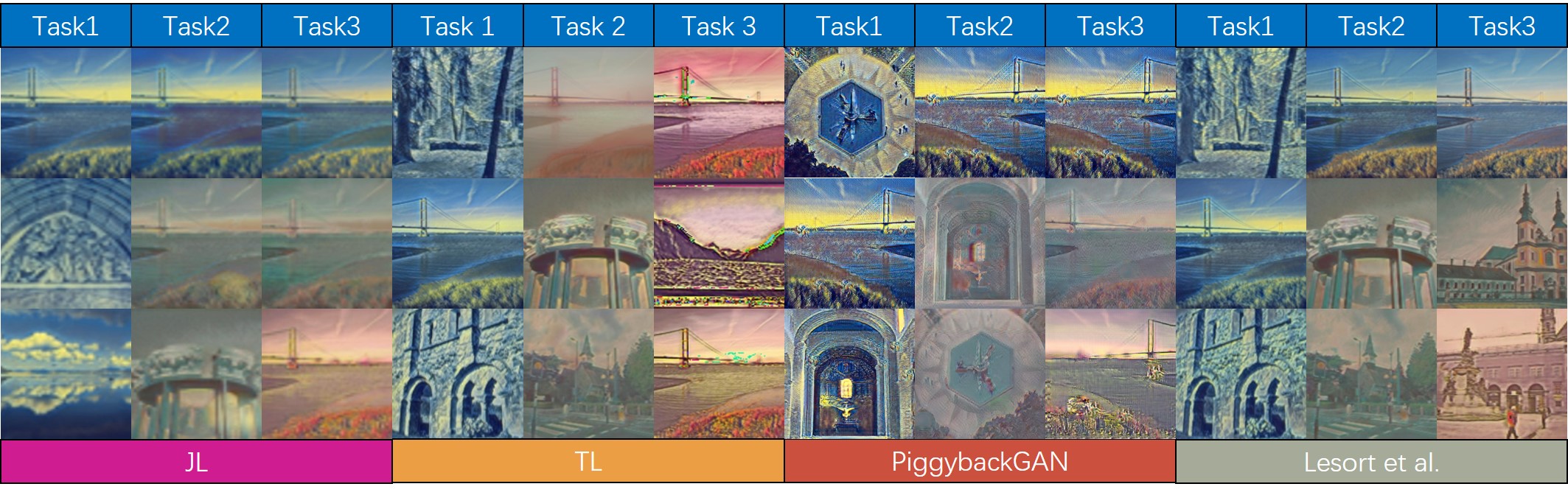}
	}\vspace{-4mm}
	\subfigure{
		\includegraphics[scale=0.3]{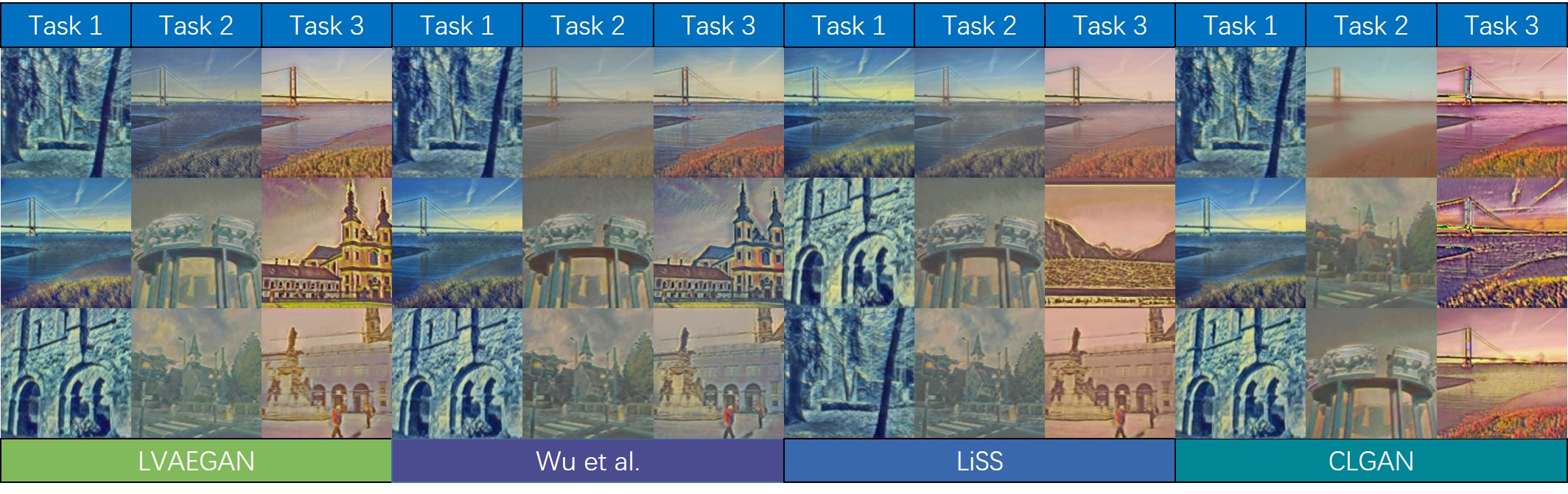}
	}\vspace{-4mm}
	\subfigure{
		\includegraphics[scale=0.3]{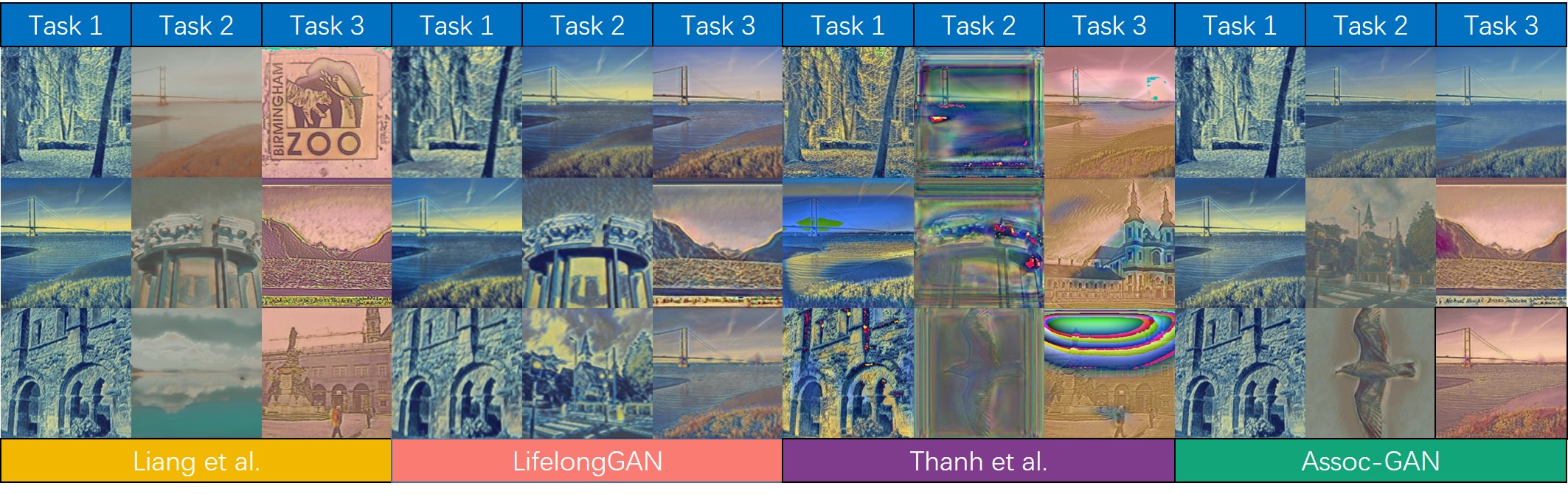}
	}
	\caption{Comparison among different methods for continual learning of image translation tasks.}	
	\label{comparison_gld}
\end{figure}
\textbf{Perceptual Study.} Next we assess the performance of Assoc-GAN on GLD style transfer tasks. We select three styles of works by Van Gogh, Monet, and Shinkai as filter templates. Each model utilizes the last task output as input domain and transfer it to the current task output domains. The transfer between arbitrary images in a given sequence is cumbersome for continual learning. Figure \ref{comparison_gld} shows some qualitative examples comparing our results with other models. JL performs better than other methods and achieves the most visually satisfactory results while TL is unable to capture previous concepts and completely forgetting the historical ones. The results of other continual learning methods are visually close to those of Assoc-GAN, indicating that both can effectively deal with catastrophic forgetting. But some reinforced generation artifacts \cite{wang2018transferring} seem exist in some historical outputs of CLGAN, PiggybackGAN and LVAEGAN. They are more sensitive to artifacts and will be reinforced during intermediate task training \cite{zhai2019lifelong}, while Assoc-GAN shows more robustness and less sensitiveness to them. To provide a reference for the quantitative analysis, we use outputs generated by \cite{johnson2016perceptual} to compare with other results. SSIM scores presented in Table \ref{table2} indicate that our model generates images that most closely resembling the painter's style and illustrates the best generalization ability among all continual methods. We can conclude that Assoc-GAN can produce qualitatively satisfactory images with more repetitive patterns compared with other methods. It also retains the content coherence and does not to be influenced by changing circumstances.

\begin{figure}
	\centering
	\subfigure{
		\includegraphics[width=0.7\linewidth]{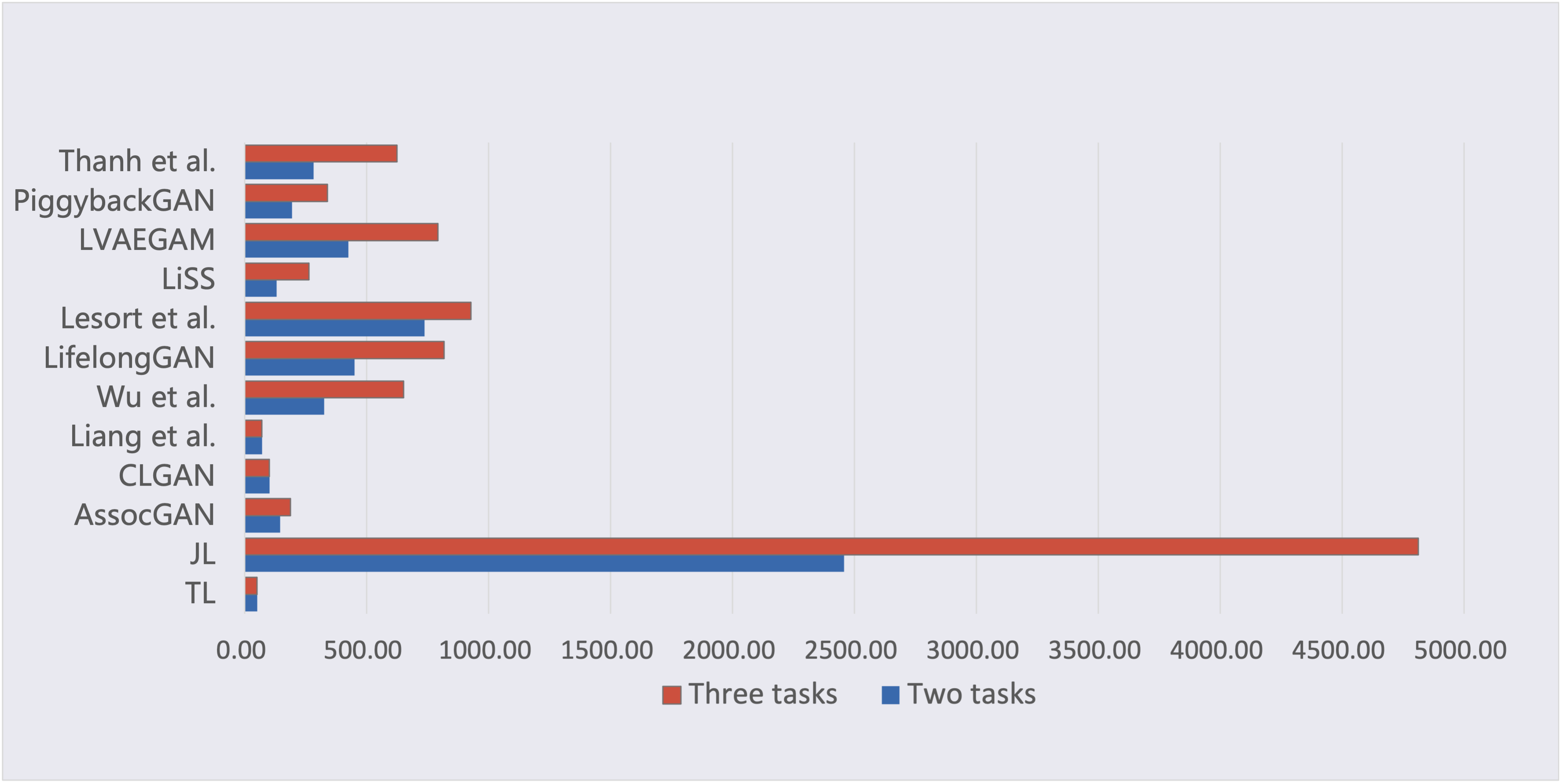}
		\label{memory_gld}
	}
	\caption{Memory study on GLD.}	
\end{figure}

\textbf{Analysis 1: Memory.} We compare the memory overhead of Assoc-GAN with some state-of-the-art approaches, as shown in Figure \ref{memory_gld}. Based on the heuristic module, an optimization that does not require access to original images, our model can achieve low spatial space overhead while obtaining promising results in comparison to memory replay methods that directly revisit past training data, such as \cite{thanh2020catastrophic}, \cite{lesort2019generative}, and JL. Compared to the above three methods, the storage on two tasks can be reduced by up to 48.4\%, 80.2\% and 94.1\%, respectively. For three tasks, our method achieves 69.8\%, 79.6\% and 96.1\% lower overhead, respectively. Although the storage of \cite{liang2018generative}, CLGAN and TL are lower than ours, their perceptual performances on GLD are the worst among all continual learning methods. Since the memory growth rates of other methods are larger than that of Assoc-GAN, it can be foreseen that continually adding tasks to the model will make the  spatial space overhead gap between other methods and Assoc-GAN larger. 

\begin{figure}
	\centering
	\subfigure{
		\includegraphics[width=0.4\linewidth]{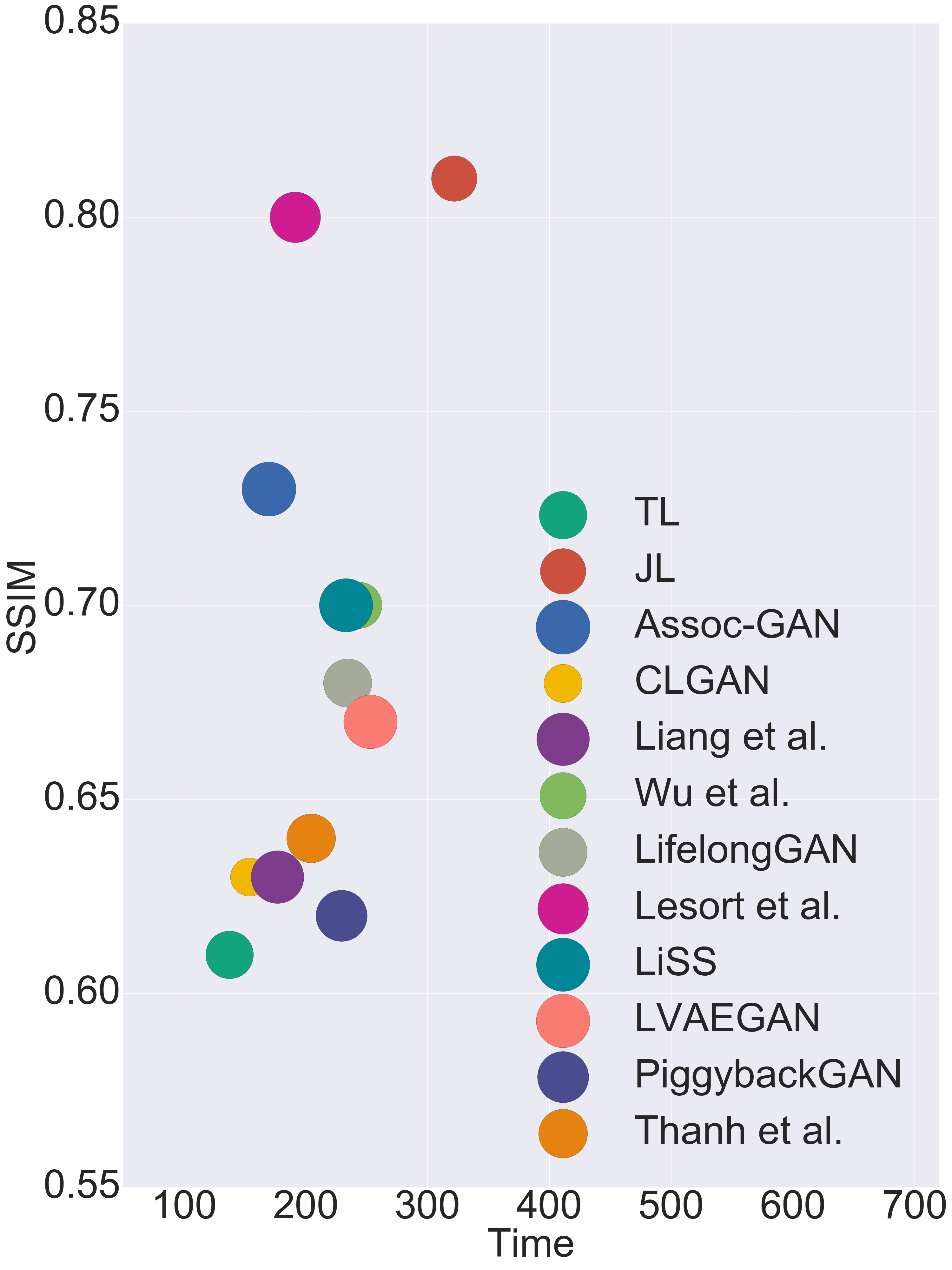}
		\label{time_gld}
	}
	\caption{Time study on GLD.}	
\end{figure}

\textbf{Analysis 2: Time.}
We present the training time and perceptual performance curve of all methods in Figure \ref{time_gld}. Note that JL sets up the upper bound for continual learning \cite{wu2018memory}, it is not subject to continual learning conditions, and it takes the achievement of the highest numerical performance for granted. TL only generates incoherent facade-like patterns and obviously completely forgets the acquired knowledge, which is the theoretical lower bound of continual learning \cite{wu2018memory}. Since our method has no access to original data or extracted feature, it utilizes the heuristics module to think of the previous circumstance from current training samples. During the training process, it only takes the time of heuristics module to backtrack the data, instead of training additional data. It greatly reduces the training time while maintaining an unforgettable generation capability compared with other continual learning methods. Assoc-GAN can reduce the time consumption by up to 27.7\% with LiSS, 40.6\% with \cite{lesort2019generative}, and 47.3\% with JL, respectively. Note that \cite{lesort2019generative} requires extra memory to store previous task data and directly revisits it in future tasks, which establishes the upper bound performance of our memory replay method \cite{abati2020conditional}. 

\begin{figure}
	\centering
	\subfigure[TL]{
		\includegraphics[scale=0.11]{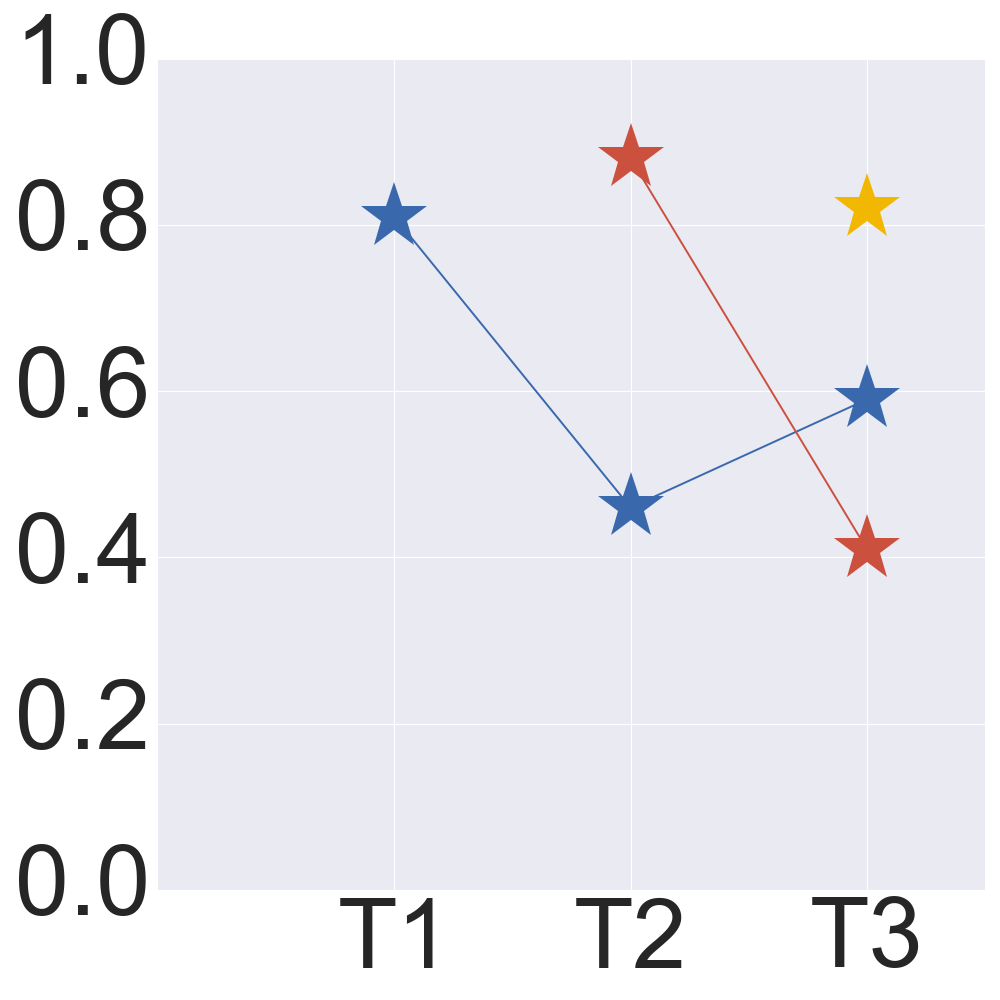}
	}\hspace{-4mm}
	\subfigure[JL]{
		\includegraphics[scale=0.11]{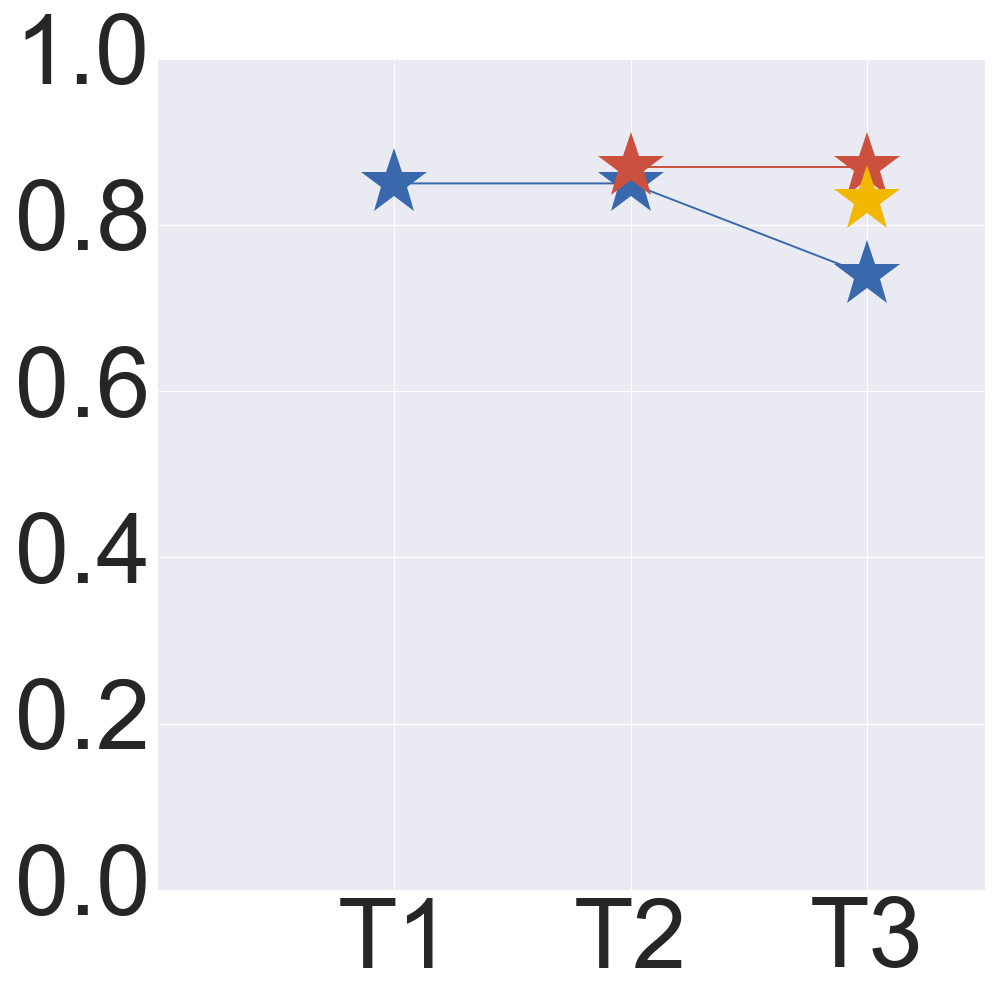}
	}\hspace{-4mm}
	\subfigure[Wu et al.]{
		\includegraphics[scale=0.11]{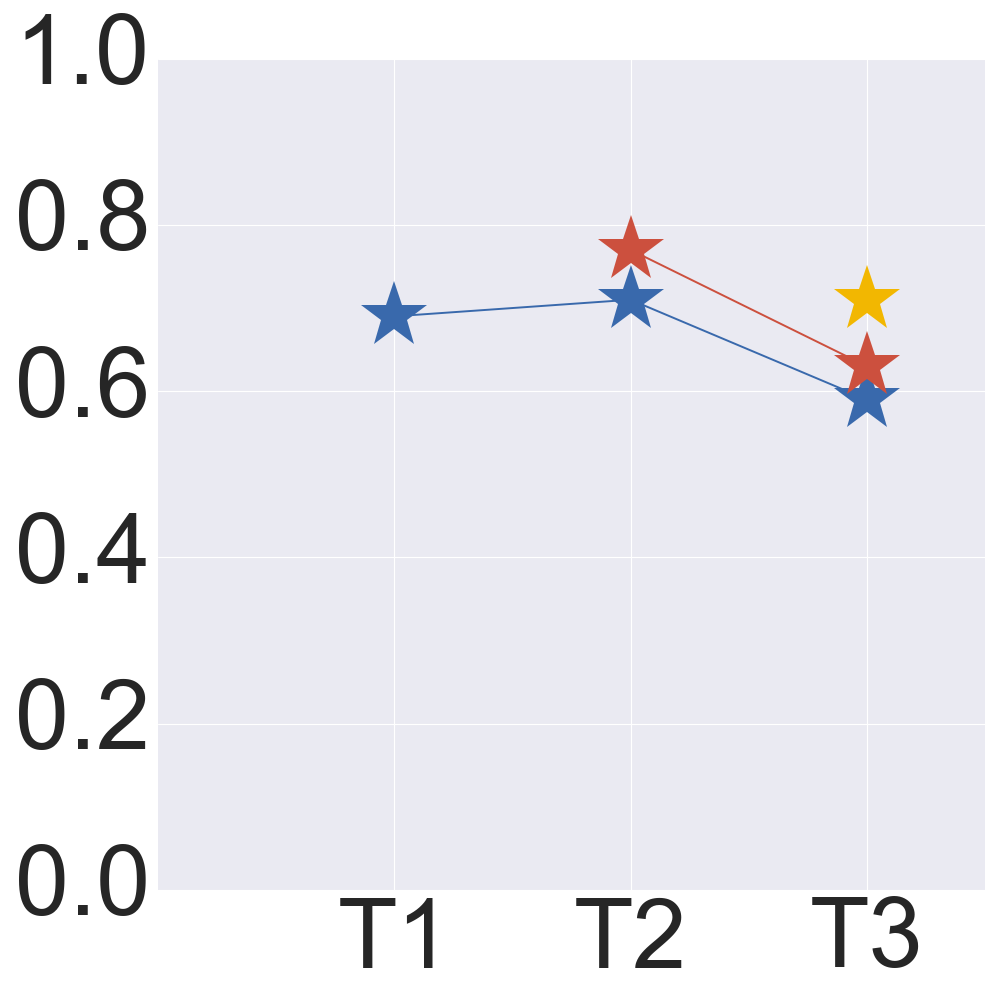}
	}\hspace{-4mm}
	\subfigure[Lesort et al.]{
		\includegraphics[scale=0.11]{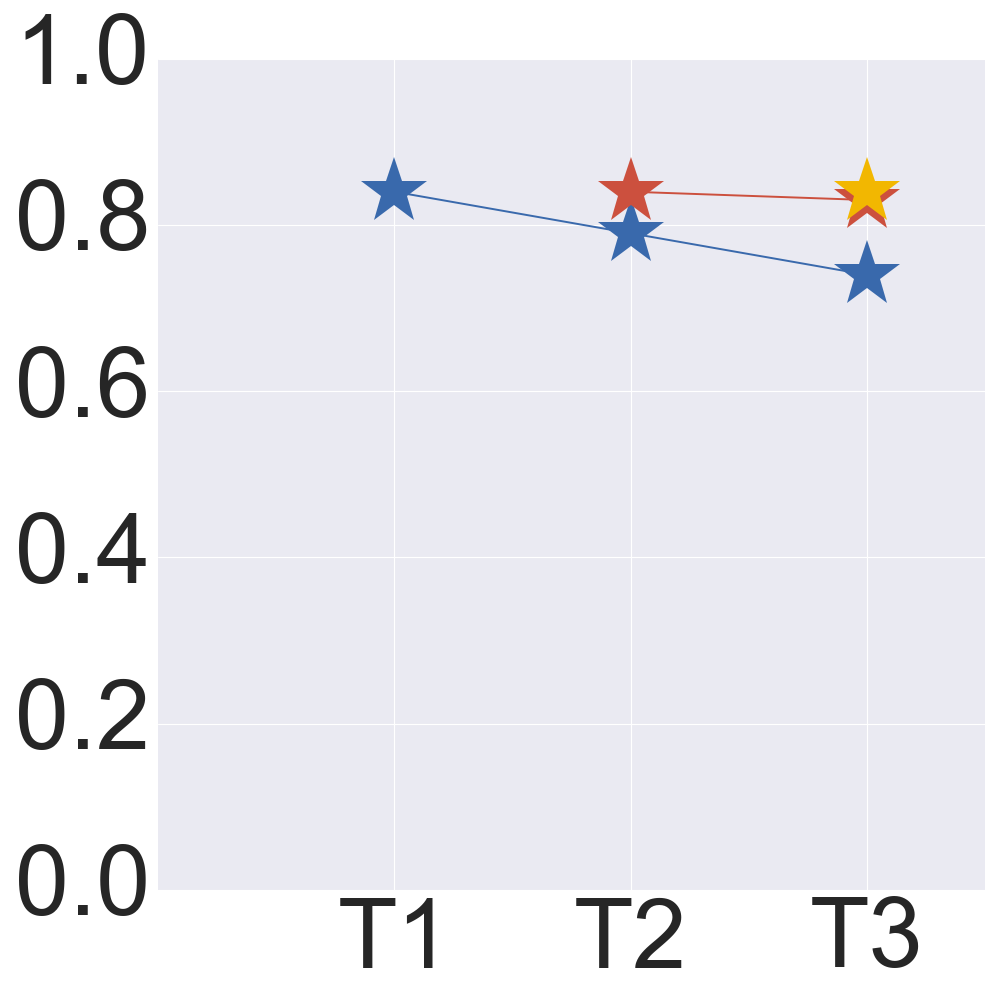}
	}\hspace{-4mm} 
	\subfigure[LVAEGAN]{
		\includegraphics[scale=0.11]{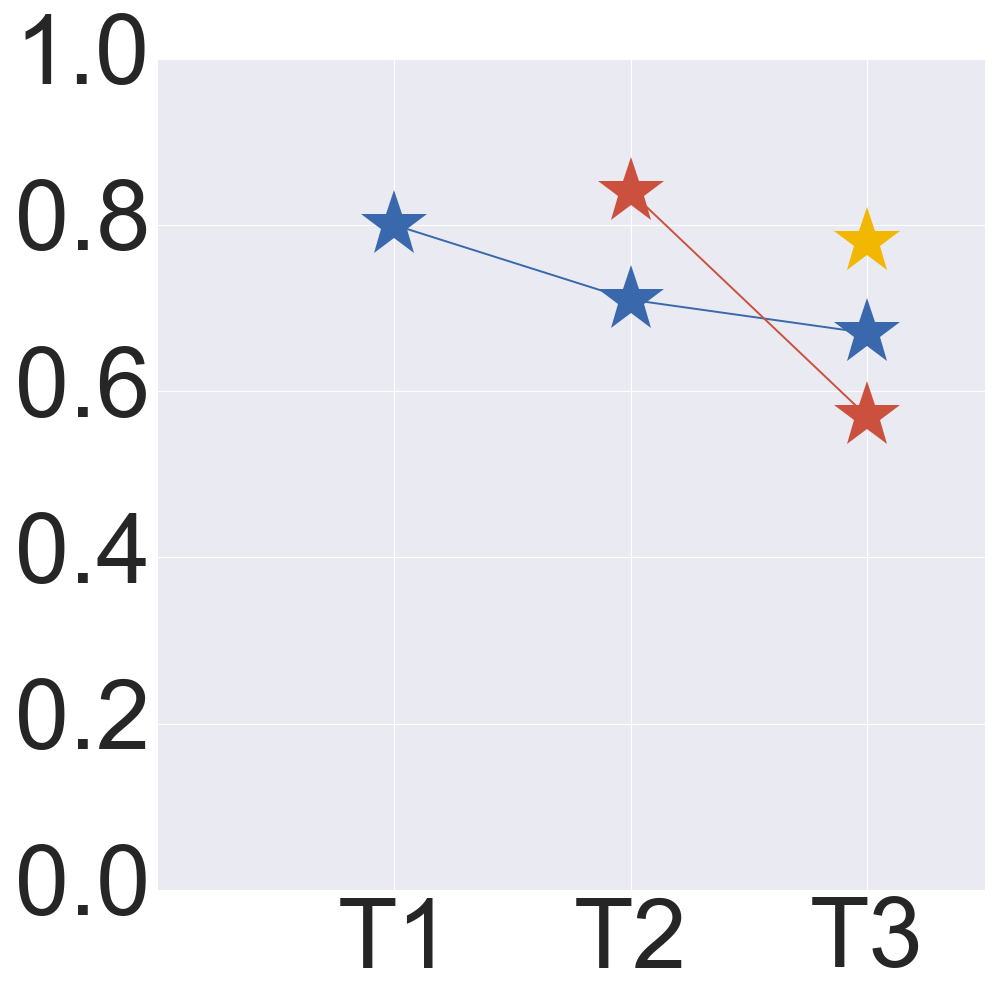}
	}\hspace{-4mm}
	\subfigure[LiSS]{
		\includegraphics[scale=0.11]{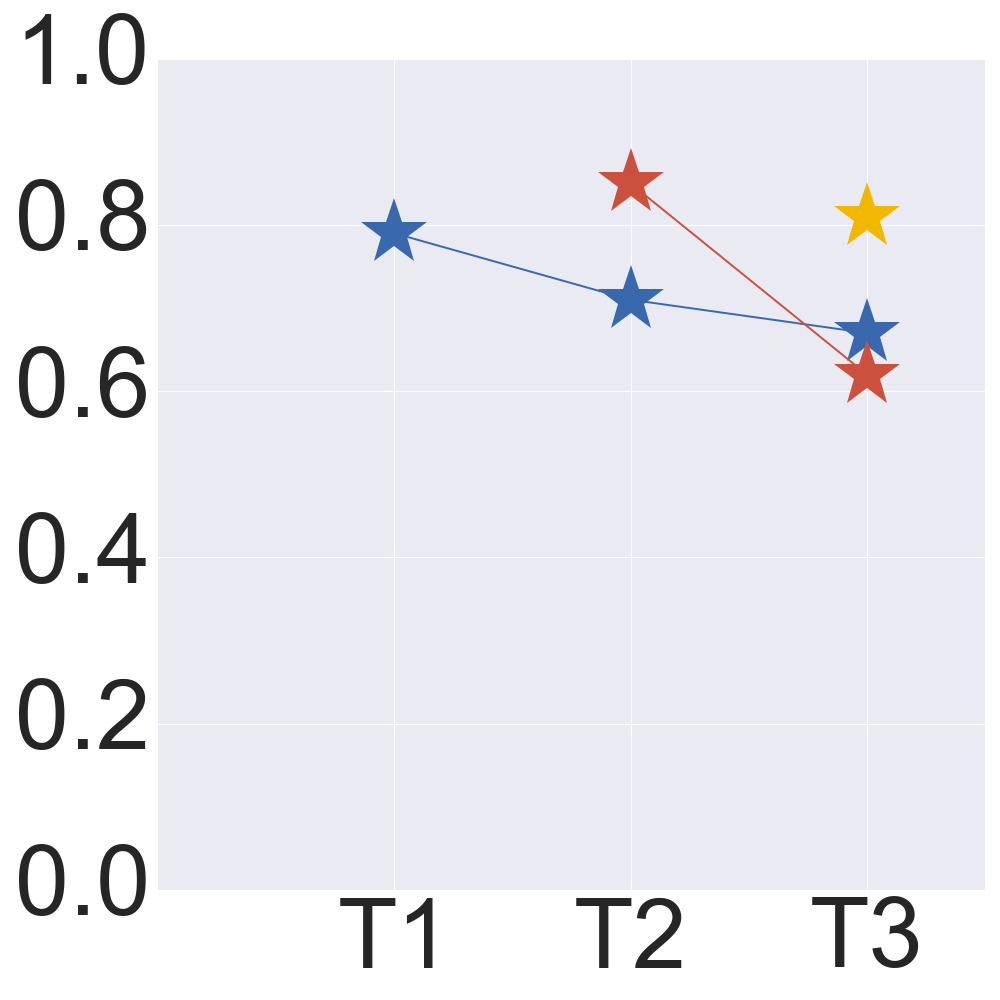}
	}\hspace{-4mm}
	\subfigure[CLGAN]{
		\includegraphics[scale=0.11]{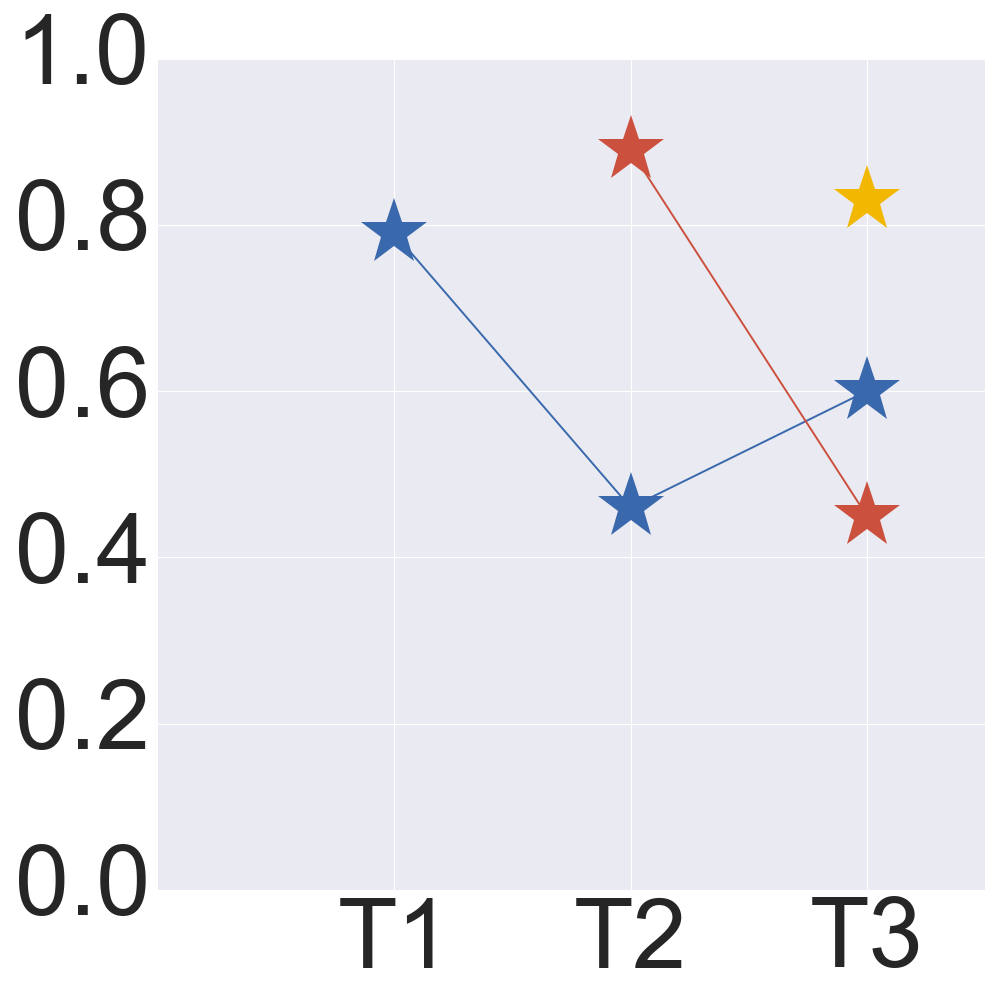}
	}\hspace{-4mm}
	\subfigure[Piggyback]{
		\includegraphics[scale=0.11]{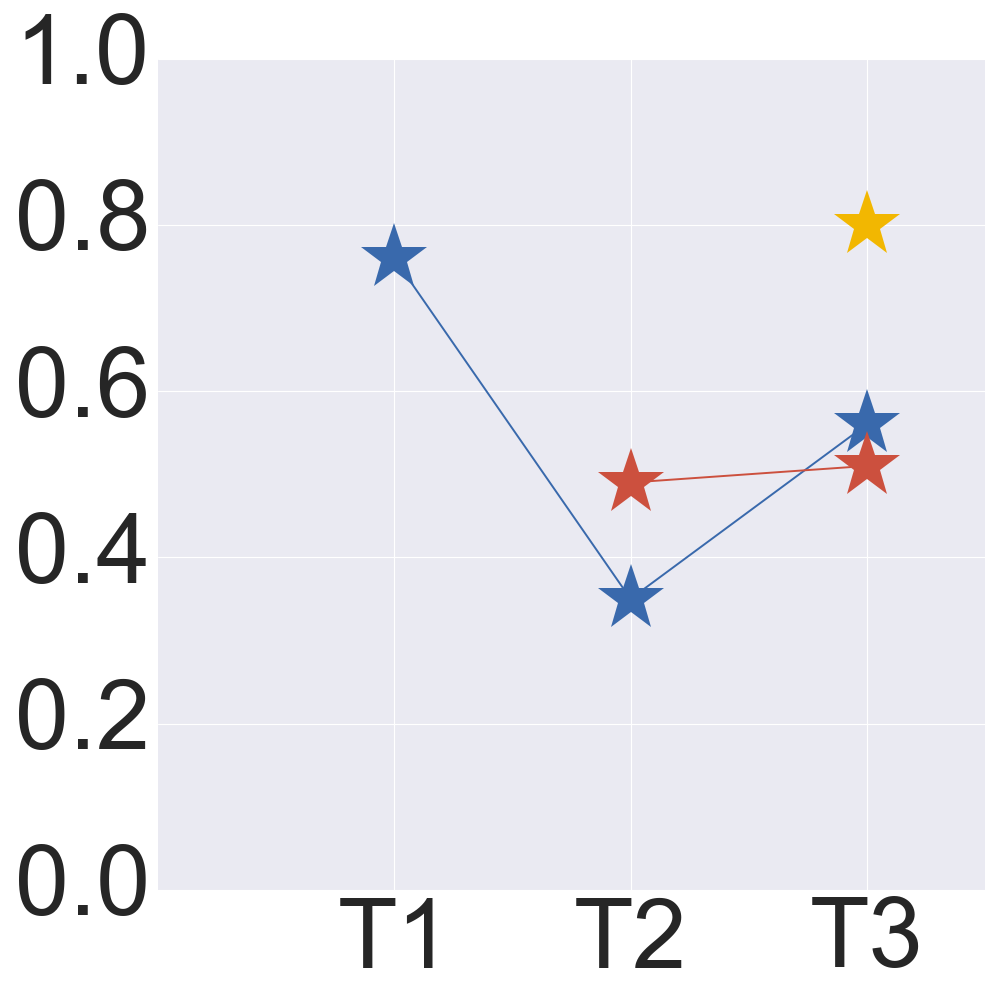}
	}\hspace{-4mm}
	\subfigure[Liang et al.]{
		\includegraphics[scale=0.11]{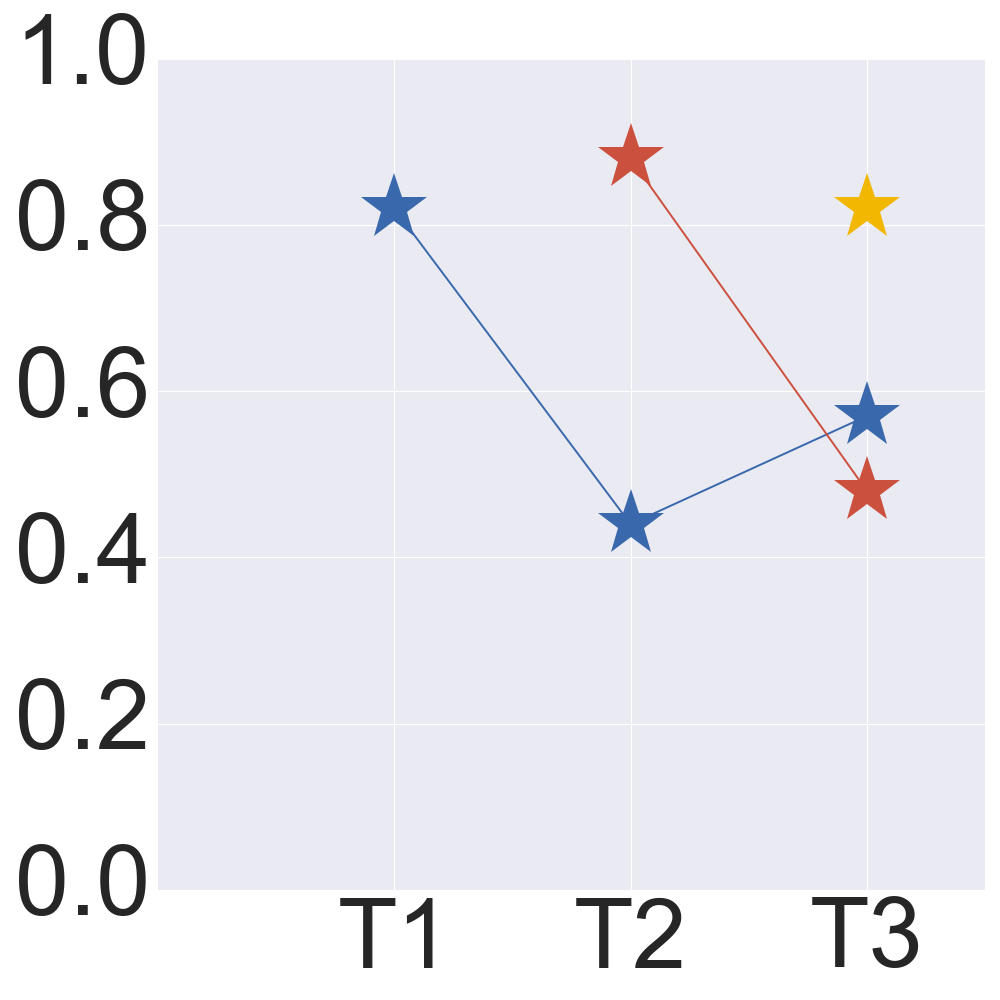}
	}\hspace{-4mm}
	\subfigure[LifelongGAN]{
		\includegraphics[scale=0.11]{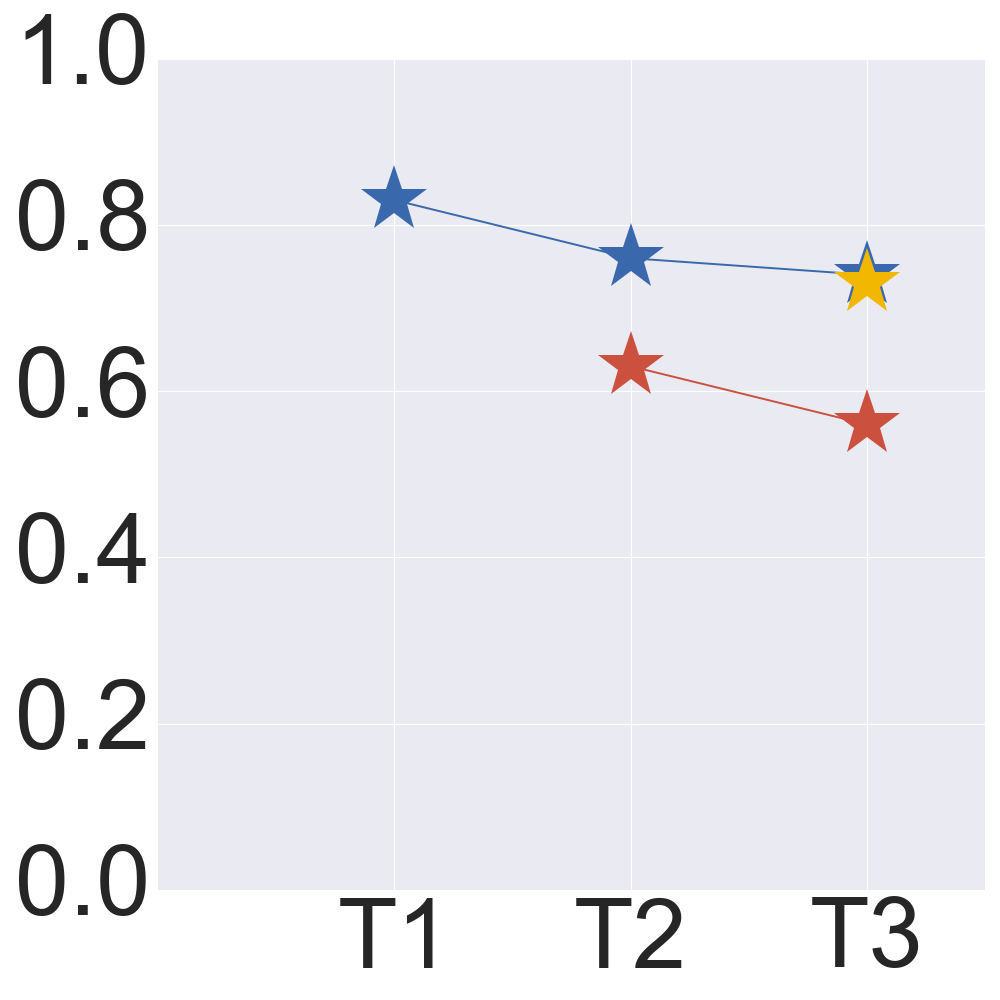}
	}\hspace{-4mm}
	\subfigure[Thanh et al.]{
		\includegraphics[scale=0.11]{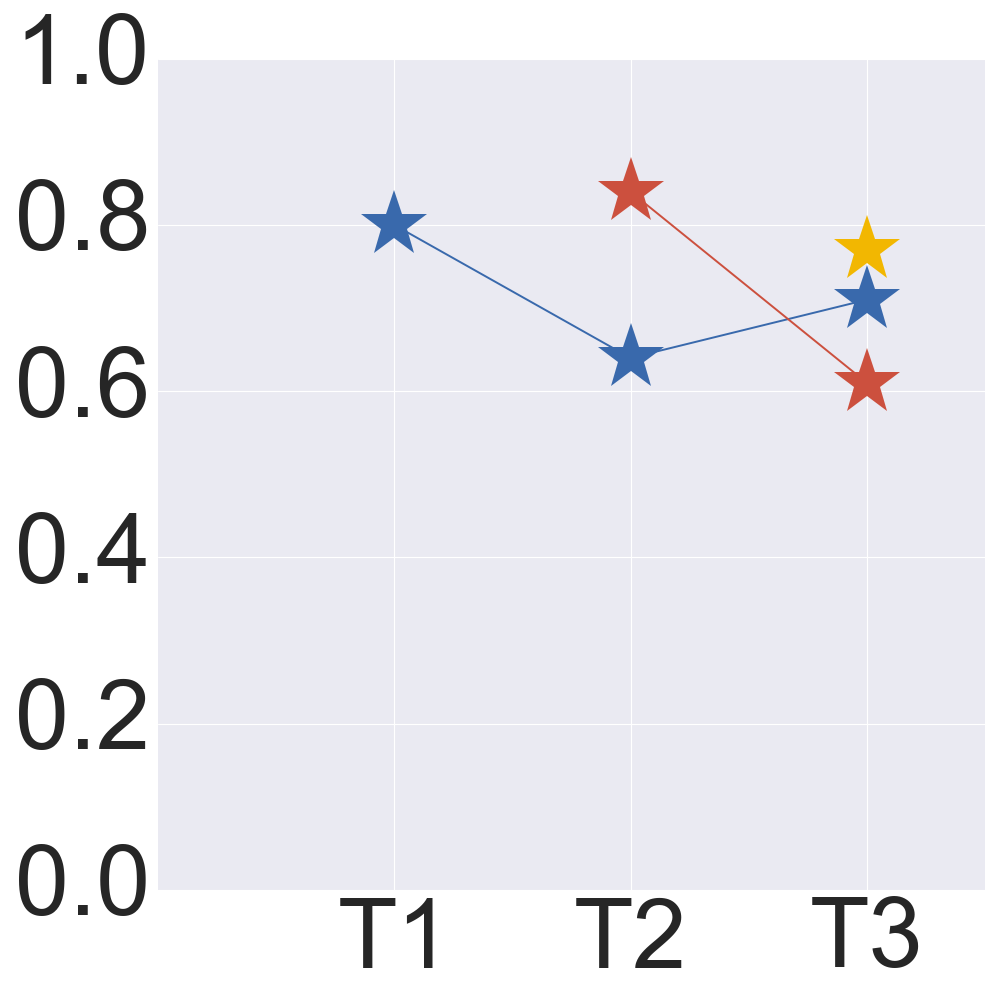}
	}\hspace{-4mm}
	\subfigure[Assoc-GAN]{
		\includegraphics[scale=0.11]{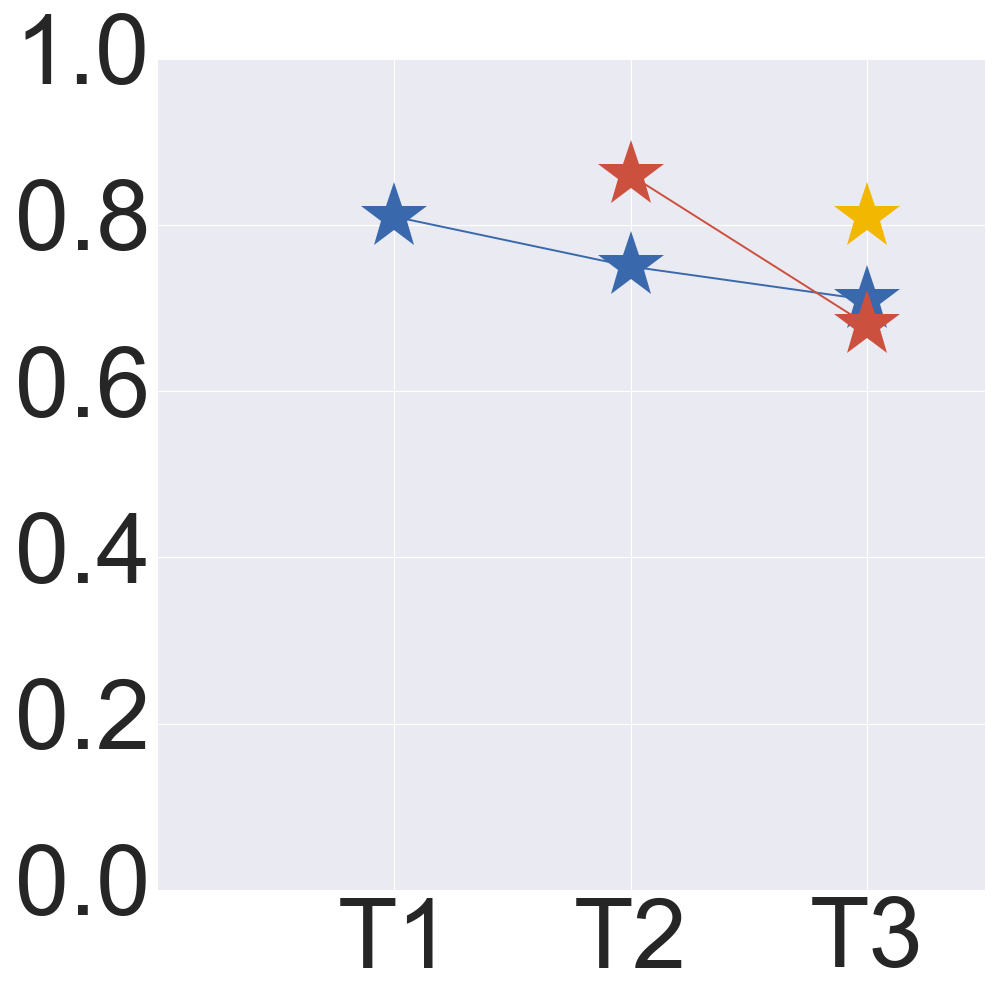}
	}
	\caption{Forgetting Analysis. The blue line is Task1, the red line is Task2, and the yellow line is Task3.}	
	\label{forget_gld}
\end{figure}

\textbf{Analysis 3: Forgetting.} 
The evolution of SSIM on three tasks for all methods during the training process is presented in the Figure \ref{forget_gld}. Since JL trains all task data together each time, it achieves the best results. TL, CLGAN, \cite{liang2018generative} and PiggybackGAN are easily to suffer from catastrophic forgetting when irrelevant tasks are fed in a sequence. In contrast to these methods, our model retains the previous task knowledge throughout sequential training on all cases. \cite{lesort2019generative}, LVAEGAN, LifelongGAN and Assoc-GAN generate competitive predictions but the knowledge forgetting speed under Assoc-GAN is slower than other three methods. It highlights the merits of our associative learning, i.e., continually updating knowledge by heuristic algorithm and generating results close to memory replay without using original data. Adding heuristics module and feature distillation significantly mitigate the catastrophic forgetting and improve the performance in all cases, especially with incremental steps over long time range.

\subsection{Further Analysis} \label{further}
\begin{figure}
	\centering
	\subfigure[Quota]{
		\includegraphics[width=0.35\linewidth]{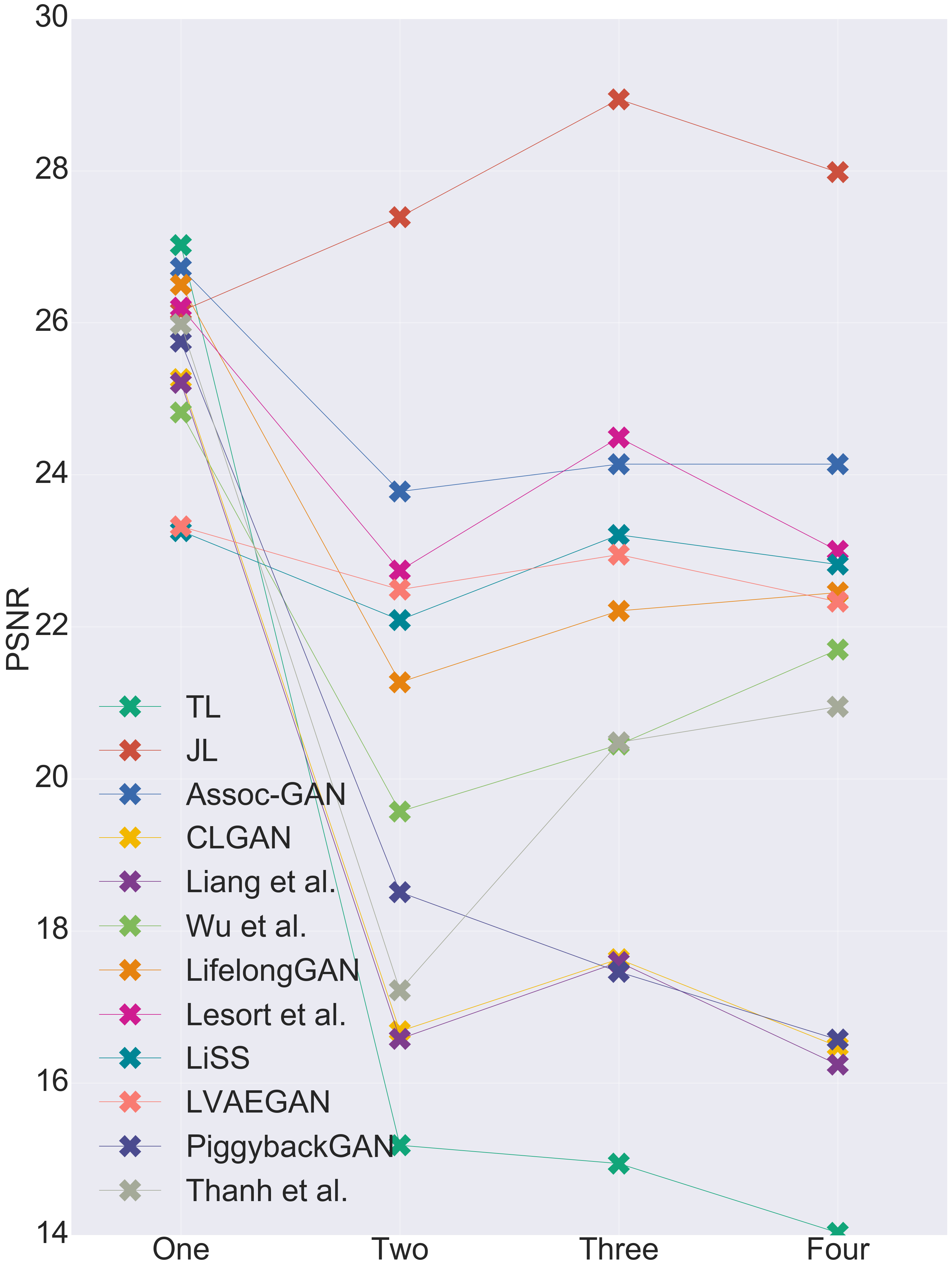}
		\label{quota}
	}\vspace{-0mm}
	\subfigure[Sequence]{
		\includegraphics[width=0.35\linewidth]{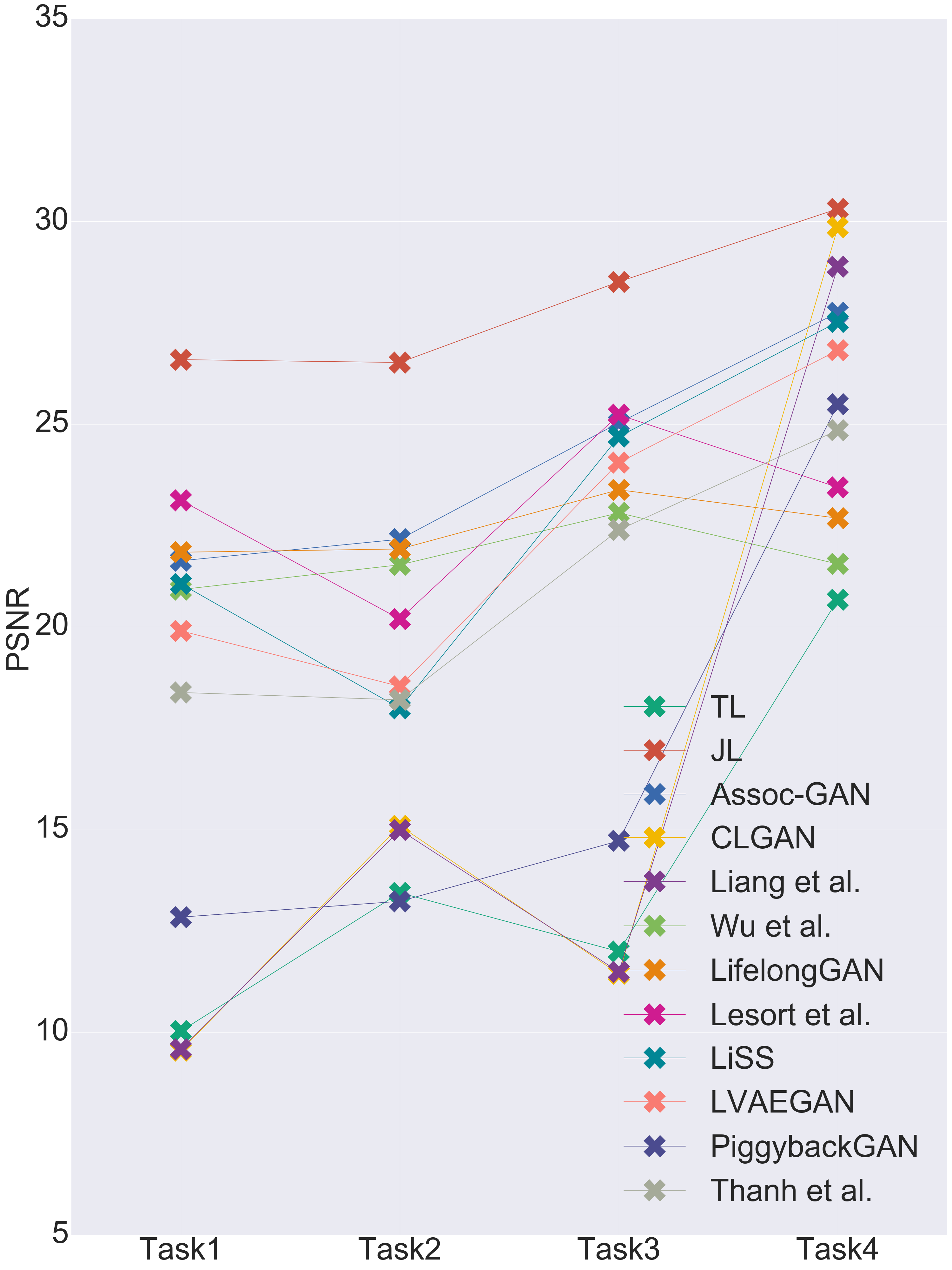}
		\label{sequence}
	}
	\caption{Quota and sequence study on DFD.}	
\end{figure}

\begin{figure}
	\centering
	\subfigure[Quota]{
		\includegraphics[width=0.35\linewidth]{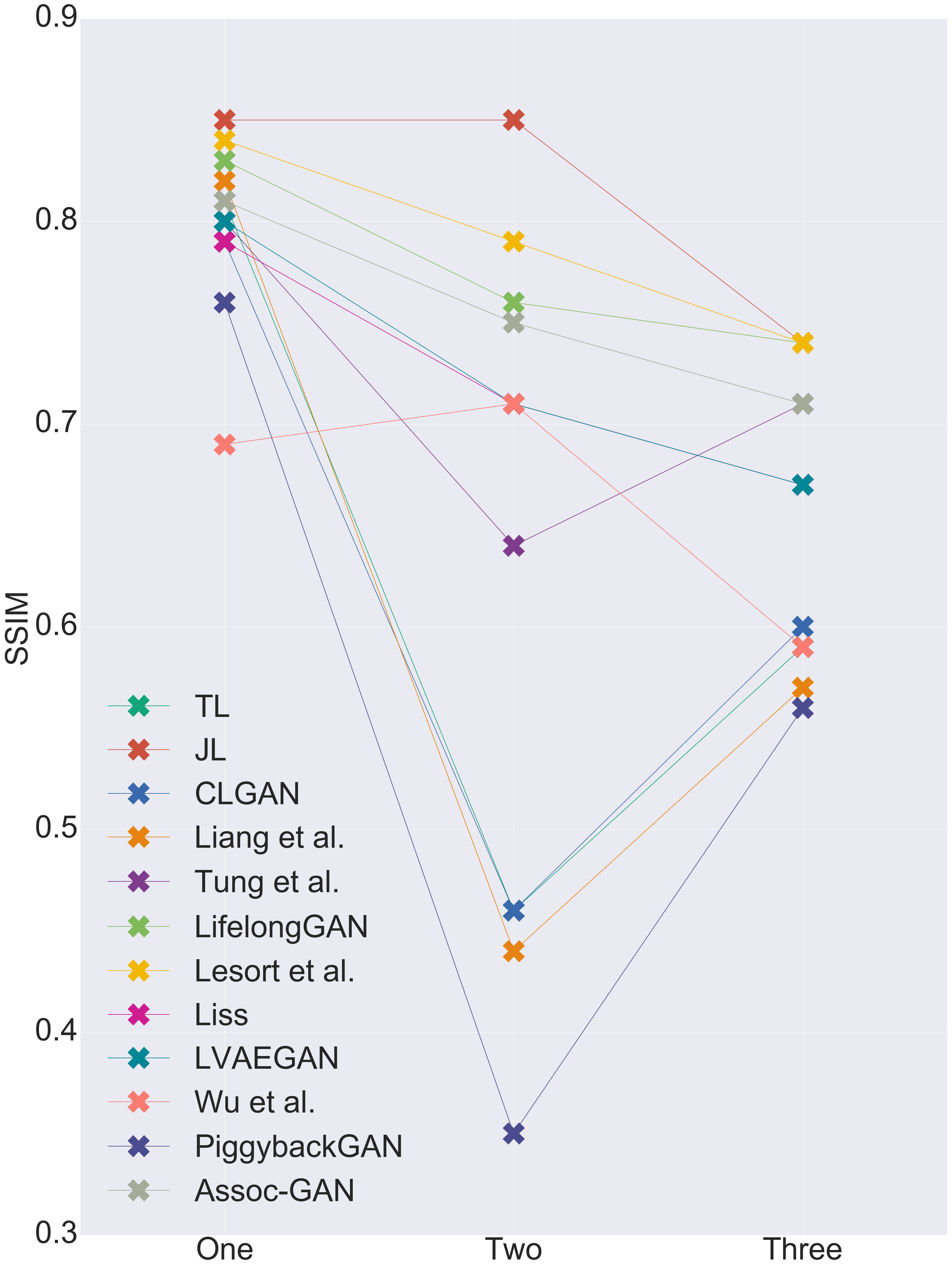}
		\label{quota2}
	}\vspace{-0mm}
	\subfigure[Sequence]{
		\includegraphics[width=0.35\linewidth]{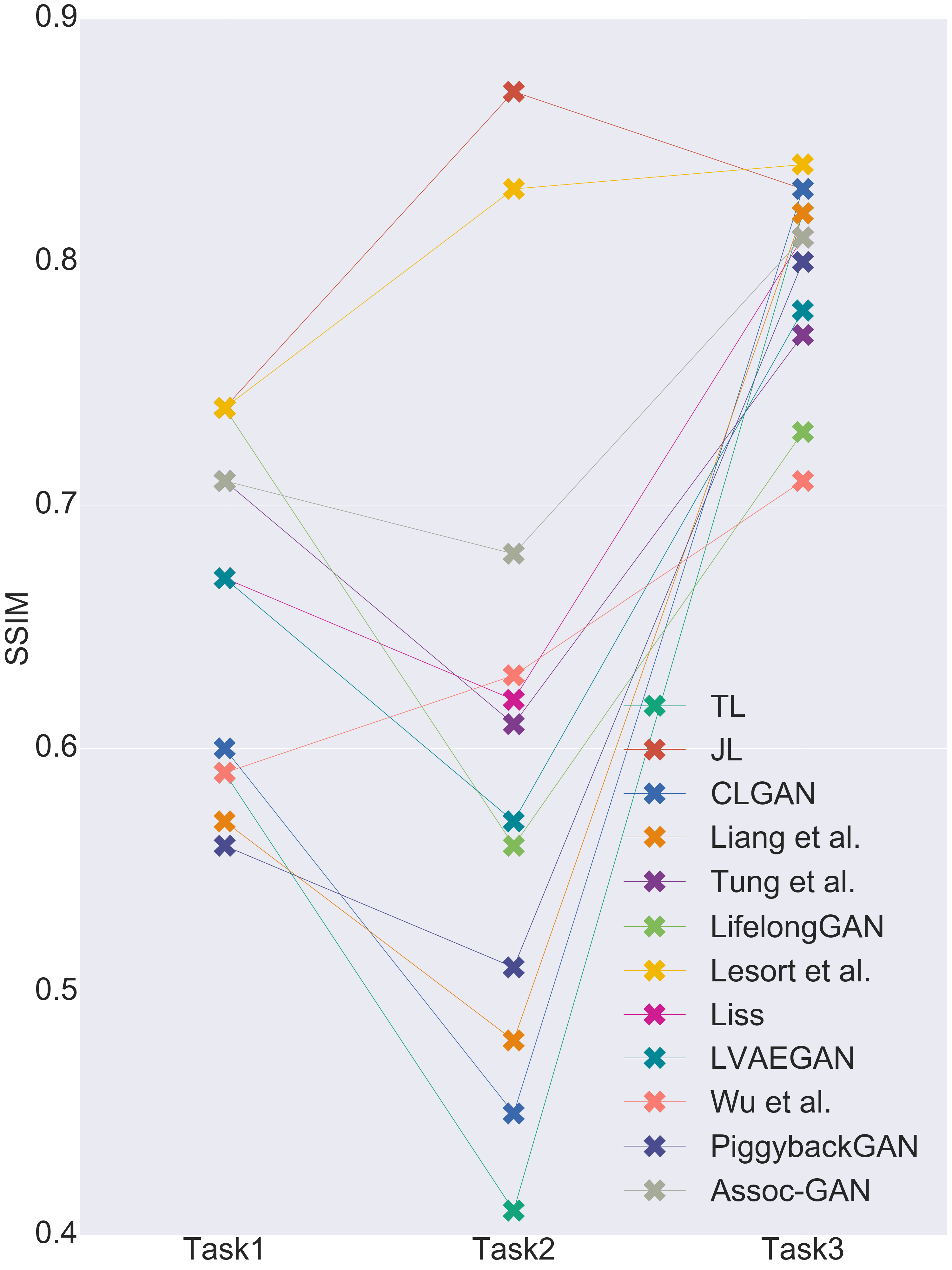}
		\label{sequence2}
	}
	\caption{Quota and sequence study on GLD.}	
\end{figure}

\textbf{Quota.} Figure \ref{quota} illustrates the average PSNR along the sequential training on DFD and Figure \ref{quota2} illustrates the average SSIM on GLD, during which new tasks are transited to the model. JL is the upper bound. TL, PiggybackGAN, CLGAN and \cite{liang2018generative} consistently achieve competitive results on the latest task while forgetting most previous tasks. The discrepancy between \cite{lesort2019generative} and Assoc-GAN is relatively small, and our model can quantitatively obtains comparable performance to it. However, since replay methods directly revisits original training samples, the perceptual quality of predictions may degrade when totally different task with inverse data distribution gradually participate in the training. The update of knowledge continually interferes with retained knowledge, and the imbalance between new and old task samples incur more fluctuations. The amplitude of \cite{lesort2019generative} is 3.46 dB, but ours is only 2.94 dB. Although the forgetting amplitude of other methods is not high, the convergence performances are far less than ours. This is because continual distillation loss intermediate the stability and plasticity, which slows down the performance decay on all incremental steps. It demonstrates that other methods are more sensitive to the quota compared with Assoc-GAN while our method can effectively reduce this bias and maintain the robustness of performance. 

\begin{figure*}[htbp]
	\centering
	\subfigure{
		\includegraphics[width=\linewidth]{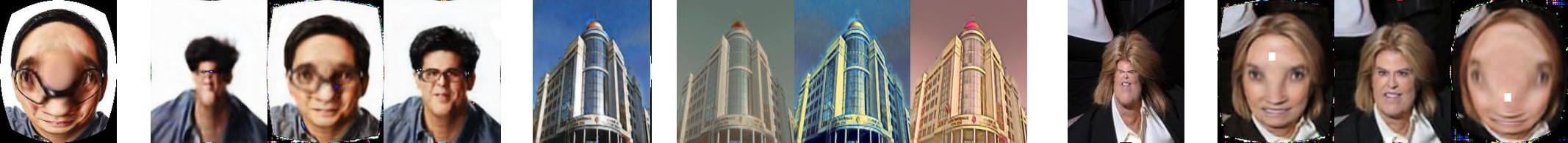}
	}\vspace{-0mm}
	\subfigure{
		\includegraphics[width=\linewidth]{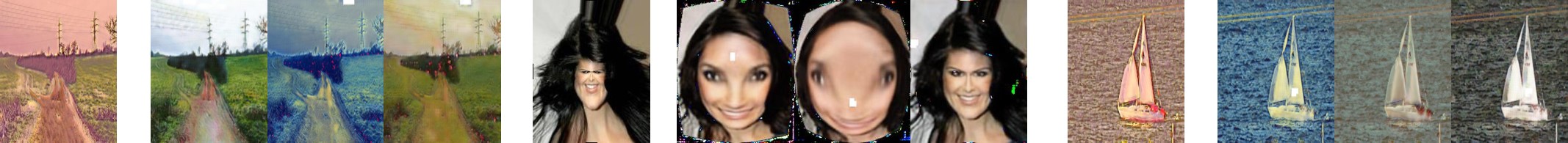}
	}\vspace{-0mm}
	\subfigure{
		\includegraphics[width=\linewidth]{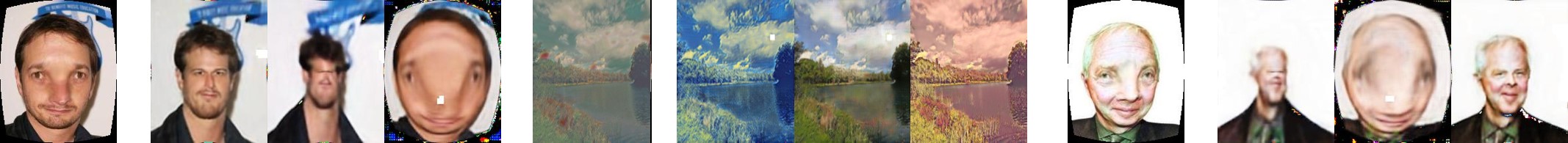}
	}\vspace{-0mm}
	\subfigure{
		\includegraphics[width=\linewidth]{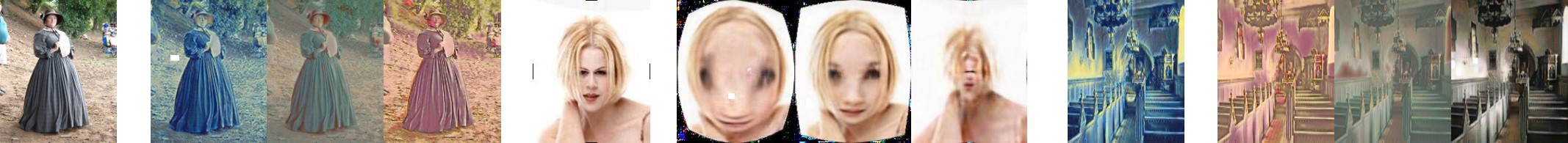}
	}
	\caption{Some examples of Association image generated by our heuristics module. The first column of each group is the original distortion image, the second to fourth column are the Associative reconstruction image.}
	\label{Assoc}
\end{figure*}
\textbf{Sequence.} Figure \ref{sequence} and Figure \ref{sequence2} demonstrate the effect of task arrival sequence on final performance and we can derive some findings on the tasks exposed to the model. In general, the more recent the task, the better the model performance. However, the performance does not systematically reduce all the time. Except for TL, PSNR of other three methods on task 2 is lower than that of task 1. \cite{lesort2019generative} drops by 12.7\% while Assoc-GAN drops by 5.8\%. In fact, task 2 and task 1 belong to different distortion types. It indicates that when new samples are totally different from what has been presented before, some new knowledge should be learned from new training samples. When they are fed into the model, the model has to adapt to the new environment and minimize the interference of previous learning, so the performance may decrease. Conversely, if new samples are similar to the previous task domain, perceptually similar distortion types or degrees will appear more frequently across different increment batches. With the assistance of knowledge accumulated by continual distillation loss, the performance of model may be improved again. For instance, PSNR of three methods on T3 all increase compared with task 2. In fact, T3 has similar distortion type with T1 but with different degree, the acquired knowledge is transferred to the training of task 3 in the form of distillation. TL does not have the ability to retain knowledge, so this law is unrepresentative to this type of method.

\textbf{Ablation Study.}
We analyze each component of Assoc-GAN to illustrate each impact on model final performance. As shown in Table \ref{table2}, UNet encounters complete failure and achieves the worst performance in all cases without the content loss. Moreover, the adversarial loss serves as an inpainting term for restoration, which is beneficial to capture the structure and encourages the network to produce sharp content. However, since tasks are reached sequentially and cannot be recurred in a long-time interval, it is necessary to utilize the heuristics module to reconstruct the current domain with large incremental steps. It illustrates that using the heuristics module improves PSNR from 14.03 dB to 21.31 dB and SSIM from 0.49 to 0.66, which contributes the most to the model performance. By adding distillation term, the two metrics are slightly improved and this is because it depresses the adaptation of new episode and guide the prediction coherent with previous circumstances. We can conclude that Assoc-GAN of exploiting continual distillation loss for the generative model proves beneficial. Ultimately, after all the components are added, our full model obtains the best results and a new state-of-art is established for continual learning.
\begin{table}
	\centering
	\scriptsize
	\resizebox{0.45\linewidth}{!}{
		\centering
		\begin{tabular}{|l|c|c|}
			\hline
			Module & PSNR & SSIM \\ \hline
			Backbone & 6.13  & 0.05  \\ \hline
			Backbone+$\mathcal{L}_{mse}$ & 13.06   & 0.46  \\ \hline
			Backbone+$\mathcal{L}_{mse}$+$\mathcal{L}_{adv}$ & 14.03  & 0.49  \\ \hline
			Backbone+$\mathcal{L}_{mse}$+$\mathcal{L}_{adv}$+$\mathcal{L}_{feature}$ & 15.87  & 0.50  \\ \hline
			Backbone+$\mathcal{L}_{mse}$+$\mathcal{L}_{adv}$+Heuristics & 21.31  & 0.66  \\ \hline
			Backbone+$\mathcal{L}_{mse}$+$\mathcal{L}_{adv}$+Heuristics+$\mathcal{L}_{feature}$ & 24.14  & 0.78  \\ \hline
		\end{tabular}
	}
	\caption{Results of each module on four tasks training.}
	\label{table2}
\end{table}

\begin{figure}
	\centering
	\subfigure[Lambda-Prior]{
		\includegraphics[width=0.4\linewidth]{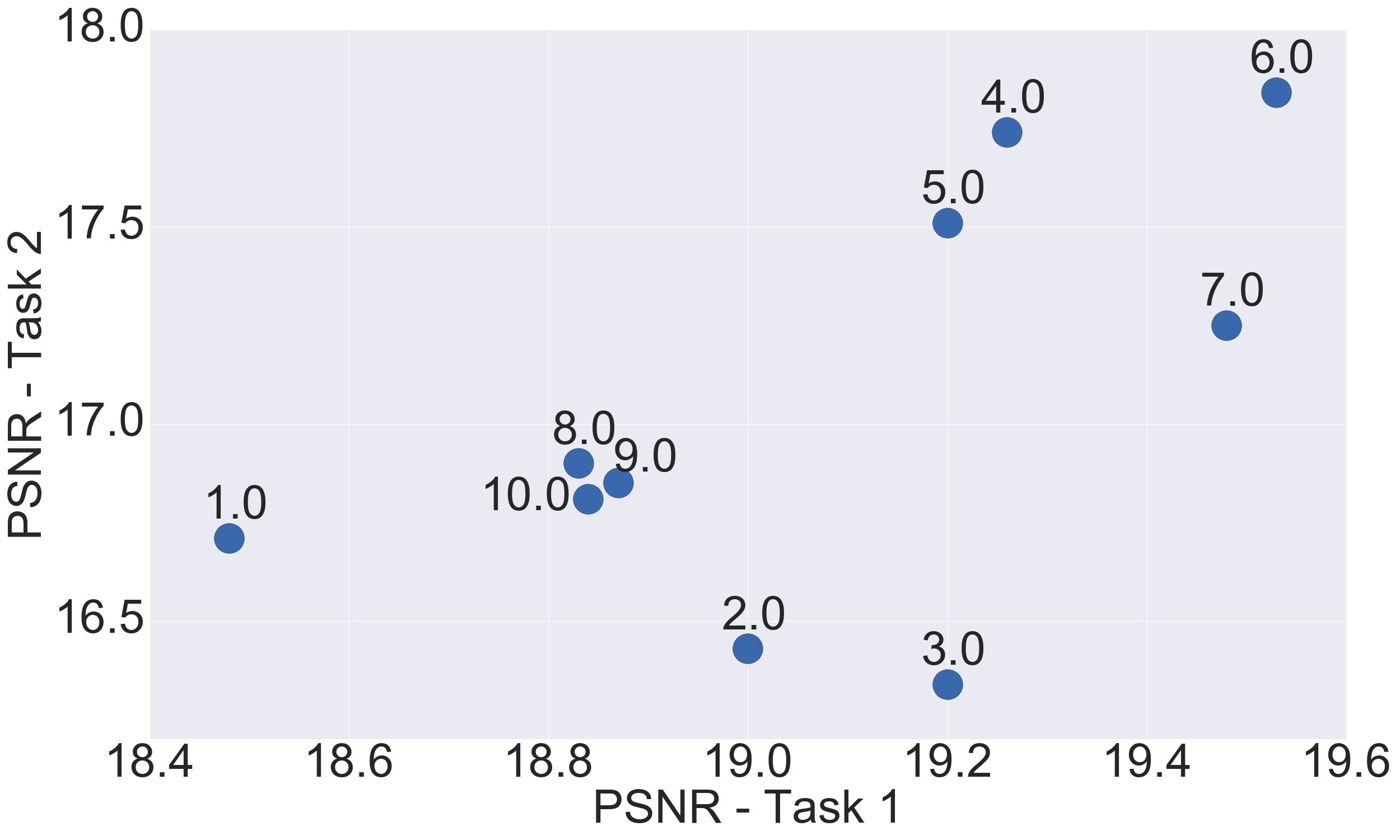}
		\label{lambda-prior1}
	}
	\subfigure[Lambda-Overall]{
		\includegraphics[width=0.4\linewidth]{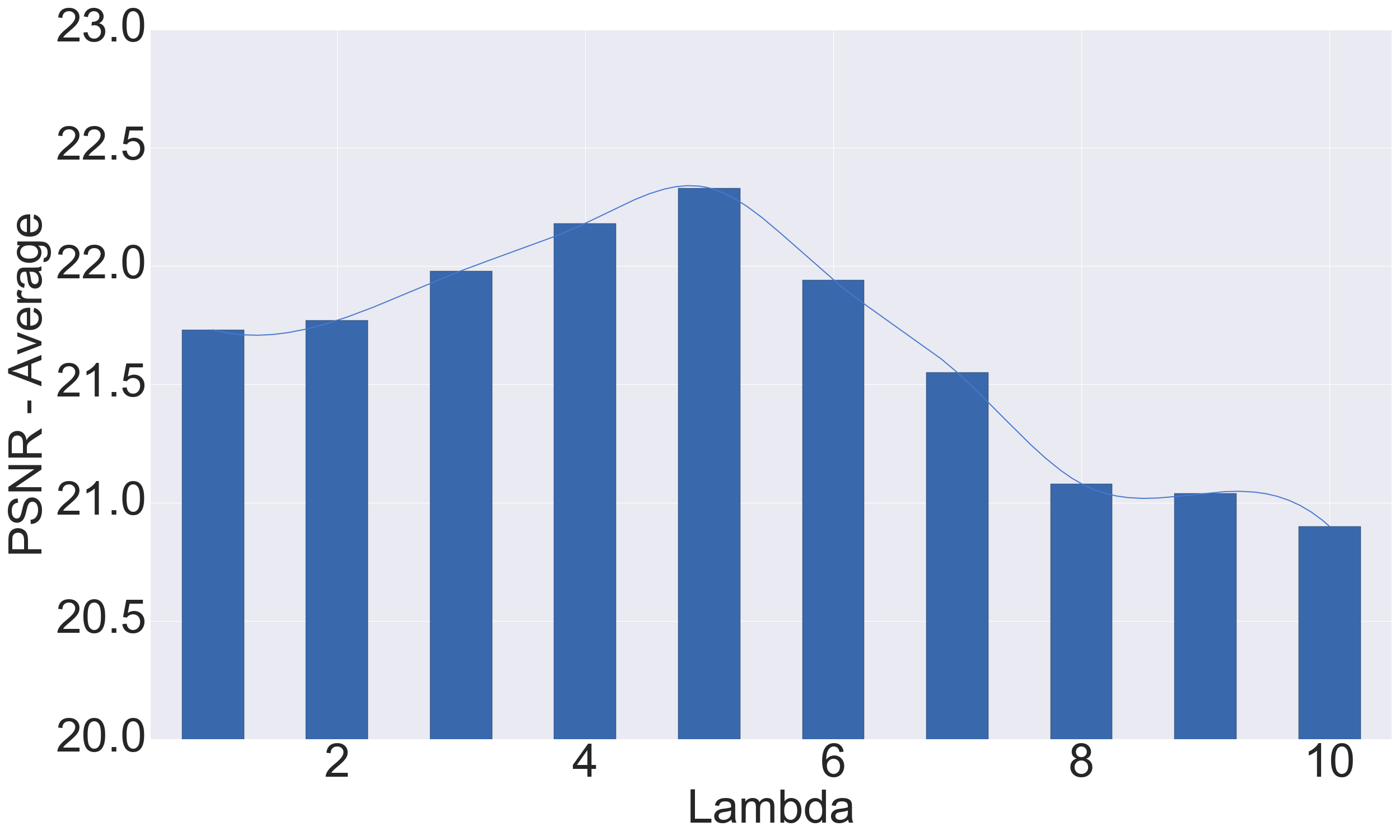}
		\label{lambda-overall1}
	}
	\caption{Hyper parameter study.}	
\end{figure}

\textbf{Hyper Parameter.} $\lambda^{'}$ is the only extra hyper parameter, which weights the balance between new and old knowledge. To understand the nature of this moderate update contributing to our solution, we adjust the $\lambda^{'}$ and observe the impact on model performance. Figure \ref{lambda-prior1} and \ref{lambda-overall1} show the impact on the average performance and the prior task performance, respectively. From Figure \ref{lambda-prior1} we can find that these scattered points can be clustered into three categories, among which 4.0, 5.0, 6.0, and 7.0 are clustered into one category, and the performance of task 1 and 2 in this category are higher than other categories. Note that as the $\lambda^{'}$ increases, more of task performance is preserved, but at some point the PSNR decays. The point where PSNR of task 1 drops represents the the optimum of the stability-plasticity curve. We also plot the final average PSNR for all the four tasks. As shown in Figure \ref{lambda-overall1}, the plateaus of PSNR indicates the balanced learning regime where the knowledge can be consolidated by updating the parameters, and $\lambda^{'}$ corresponding to it is 5.0. We identify the important hyper parameter to update knowledge where our model can confidently learn new tasks while still accumulating knowledge over time.

\textbf{User Study.} We conduct a user study with 20 participants to estimate the perception of generated images. We show each participant 60 groups of pictures, each group contains our generated predictions and those of other methods. Participants should point out the most realistic pictures in each group. Table \ref{study} presents the percentage of votes attained by each method in each task. This study indicates that our approach is close to direct memory replay but not par with JL. Since JL is the theoretical upper bound of continual learning, it achieves the best visual results and makes participants more satisfied in perception. Compared with \cite{lesort2019generative}, which is also a continual learning method, our performance is still very close to it. Despite this, our method has achieved satisfactory results to a certain extent, with a satisfaction rate of approximately 25\%. TL does not have the capability for continual learning, its results of all previous tasks can not make participants feel visually satisfied except the most recent task. 
\begin{table}[h]
	\centering
	\scriptsize
	\resizebox{0.45\linewidth}{!}{
		\centering
		\begin{tabular}{|c|cccc|}
			\hline
			& T1 & T2 & T3 & T4 \\ \hline
			JL & 0.53  & 0.45  & 0.38  & 0.33  \\ 
			TL & 0.00  & 0.00  & 0.00  & 0.23  \\ 
			\cite{lesort2019generative} & 0.18  & 0.27  & 0.33  & 0.20  \\ 
			Assoc-GAN & 0.28  & 0.22  & 0.28  & 0.23  \\ \hline
	\end{tabular}}
	\label{study}
	\caption{User Study.}
\end{table}

\section{Conclusion}
In this paper, we explore the catastrophic forgetting problem for generative network and propose a brain-inspired associative learning framework based on the mediation by potentiative heuristics module and depressive continual distillation. We evaluate Assoc-GAN and other generative models on image translation task. It demonstrates that Assoc-GAN optimizes the generation performance in the following aspects: 1) generating comparable results to others while not forgetting historical tasks; 2) maintaining the same performance with others while reducing spatial space overhead by up to 91.0\%; 3) The training time is dropped by up to 69.39\%.

\clearpage
\newpage

\appendix
\section{Associative Learning Mechanism}
As shown in \ref{brain}, the formation of association depends on the dopaminergic reinforcement mediated by two dopamine receptors, DopR1 and DopR2. These two signaling cascades exhibit different temporal sensitivity, instructing opposing influences on associative learning.

As discussed in Section \ref{introduction}, human can heuristically associate one picture with another relative picture. This is due to the nature of neuroscience that the formation and update of association are mediated by potentiation and depression stimulation. As shown in the associative learning model in Figure \ref{brain}, every node represents a neuron, and red nodes refer to neurons that are essential for associative learning. Normally, primitive and stable connections exist between neurons, which are represented by full lines. If two dopamine receptors, DopR1 and DopR2, stimulate almost at the same time and repeatedly in the network, the original network will be trained into an associative network \cite{handler2019distinct}. To be more specific, backward pairing of DopR1 and DopR2 ensures the potentiation, and weak connections between neurons, which are represented by dotted lines, will be potentiated to stable connections. Otherwise, the forward pairing of DopR1 and DopR2 induces the depression, and weak connections will be depressed to cut off connections. Word fluency test, which requires examinees to write words beginning with certain letter, is used to quantify the capability of association.
\section{Theoretical Analysis}
In the main paper, we provide distillation loss of Assoc-GAN under continual learning situation. Here we give detailed proof for our choice of constraint.

Given a data set $\mathcal{D}$, our goal is to find the most likely solution $\theta$ to fit the distribution of data $\rightarrow$ task, by calculating the conditional probability $p(\theta|\mathcal{D})$.

\begin{small}
	\begin{equation}
		\begin{aligned}
			p(\theta|\mathcal{D}) &= \frac{p(\theta , \mathcal{D})}{p(\mathcal{D})}\\
			&= \frac{p(\mathcal{D}|\theta)p(\theta)}{p(\mathcal{D})}
		\end{aligned}
	\end{equation}
\end{small}

then we use $\log$ function in the left and right, simultaneously. We can get the equation as follows.

\begin{small}
	\begin{equation}
		\log p(\theta|\mathcal{X},\mathcal{Y})=\log p(\mathcal{X},\mathcal{Y}|\theta)+\log p(\theta) -\log p(\mathcal{X},\mathcal{Y}),
		\label{joint}
	\end{equation}
\end{small}where $p(\mathcal{X},\mathcal{Y})$ is the probability of data $\mathcal{D}$ and $\log p(\theta)$ is the prior probability of parameters that can match all tasks.

For continual learning, all task data is broken up into $n$ batches $\mathcal{D}=\{\mathcal{D}_{1},...,\mathcal{D}_{n}\}$.

According to the equation(1), we can find that:

\begin{small}
	\begin{equation}
		\begin{aligned}
			&p(\theta_{n}|\mathcal{D}) = p(\theta_{n}|\mathcal{D}_{1},...,\mathcal{D}_{n})\\
			&=\frac{p(\mathcal{D}_{1},...,\mathcal{D}_{n},\theta_{n})}
			{p(\mathcal{D}_{1},...,\mathcal{D}_{n})}\\
			&=\frac{p(\mathcal{D}_{1})p(\mathcal{D}_{2}|\mathcal{D}_{1})...p(\theta_{n}|\mathcal{D}_{1},...,\mathcal{D}_{n-1})
				p(\mathcal{D}_{n}|\mathcal{D}_{1},...,\mathcal{D}_{n-1},\theta_{n})}
			{p(\mathcal{D}_{1})p(\mathcal{D}_{2}|\mathcal{D}_{1})...p(\mathcal{D}_{n}|\mathcal{D}_{1},...,\mathcal{D}_{n-1})}\\
			&=\frac{p(\theta_{n}|\mathcal{D}_{1},...,\mathcal{D}_{n-1})
				p(\mathcal{D}_{n}|\mathcal{D}_{1},...,\mathcal{D}_{n-1},\theta_{n})}
			{p(\mathcal{D}_{n}|\mathcal{D}_{1},...,\mathcal{D}_{n-1})}
		\end{aligned}
	\end{equation}
\end{small}

Because ${\mathcal{D}_{1},...,\mathcal{D}_{n}}$ are independent distribution, so we can get an equation as follows.

\begin{small}
	\begin{equation}
		p(\theta_{n}|\mathcal{D})=\frac{p(\theta_{n}|\mathcal{D}_{1},...,\mathcal{D}_{n-1})p(\mathcal{D}_{n}|\theta_{n})}{p(\mathcal{D}_{n})}
	\end{equation}
\end{small}

Finally, we also use $\log$ function in both sides. we can get the equations as follows.

\begin{small}
	\begin{equation}
		\begin{aligned}
			\log p(\theta_{n}|\mathcal{D})&=\log p(\mathcal{D}_{n}|\theta_{n})-\log p(\mathcal{D}_{n}) +
			\log p(\theta_{n}|\mathcal{D}_{1},...,\mathcal{D}_{n-1})\\
			&=\log p(\mathcal{D}_{n}|\theta_{n})-\log p(\mathcal{D}_{n})+\log p(\theta_{n}|\mathcal{D}_{past})
		\end{aligned}
	\end{equation}
\end{small}

Then we can get the equation on the paper as follows.

\begin{small}
	\begin{equation}
		\begin{aligned}
			\log p(\theta_{i}|\mathcal{X},\mathcal{Y})=&\log p(\mathcal{X}_{i},\mathcal{Y}_{i}|\theta_{i}) -\log p(\mathcal{X}_{i},\mathcal{Y}_{i}) \\& +\log p(\theta_{i}|\mathcal{X}_{past},\mathcal{Y}_{past})
		\end{aligned}
	\end{equation}
\end{small}

where $\{\mathcal{X}_{i},\mathcal{Y}_{i}\}$ is the data distribution of the i-th task, and $\{\mathcal{X}_{past},\mathcal{Y}_{past}\}$ is the data distribution of the previous i-1 tasks. Note that the probability of i-th task data $p(\mathcal{X}_{i},\mathcal{Y}_{i})$ is a constant value, and the log probability of i-th task data given i-th task parameters $\log p(\mathcal{X}_{i},\mathcal{Y}_{i}|\theta_{i})$ is simply tantamount to the preliminary loss function at hand $\mathcal{L}$. 

$\log p(\theta_{i}|\mathcal{X}_{past},\mathcal{Y}_{past})$ is a posterior probability distribution term and it contains information about all previous task data, which is difficult to calculate.
In order to get an approximation of the probability $\log p(\theta_{i}|\mathcal{X}_{past},\mathcal{Y}_{past})$, we can use the Laplace Approximation method \cite{mackay1992practical}. 

Let $f(\theta)$ be the probability density function $pdf$ of $p(\theta|\mathcal{X}_{past},\mathcal{Y}_{past})$ as known as $ p(\theta|D_{past})$.

According to \cite{cevherlaplace}, we can using the Laplace approximation to approximate a single-mode $p d f$ with a Gaussian:
\begin{enumerate}
	\item find a local maximum $\theta^*$ of the given $p d f f(\theta)$
	\item calculate the variance $\sigma^{2}=-\frac{1}{f^{\prime\prime}\left(\theta^*\right)}$
	\item approximate the $p d f$ with $\log p(\theta|D_{past}) \approx \mathcal{N}\left[\theta^*, \sigma^{2}\right]$
\end{enumerate}

$\theta*$ in the network is a vector of parameters considered as $(\theta_1,\theta_2 ... \theta_n)$, so $f^{\prime \prime}\left(\theta^*\right)$ is a Hessian Matrix, which is $f^{\prime \prime}\left(\theta^*\right) = (\frac{\partial^{2} f(\theta)}{\partial^{2} \theta}|_{\theta=\theta^{*}})$.

First step, we need to find a local maximum $\theta^*_{past}$ of the past works. In the past tasks, we can get $\log p(D_{past} \mid \theta_{past}^{*})$ by training the network which is considered as a process of the Maximum Likelihood Estimation(MLE).

\begin{small}
	\begin{equation}
		\begin{aligned}
			\log p(D_{past} \mid \theta_{past}^{*}) &= \log p(\theta_{past}^{*}|D_{past})\\ 
			&+\log p(D_{past})- \log p(\theta_{past}^{*})
		\end{aligned}
	\end{equation}
\end{small}

Because $\log p(\theta_{past}^{*})$ and $\log p(D_{past})$ were a constant value, $\log p(\theta_{past}^{*}|D_{past})$ can be the local maximum, which means $\log f(\theta)$ can have the local maximum at $\theta^*_{past}$.

Second step, we calculate the variance $\sigma^{2}=-\frac{1}{f^{\prime \prime}\left(\theta^*_{past}\right)}$. $f^{\prime \prime}\left(\theta^*_{past}\right)$ is a Hessian Matrix. Due to the complexity of calculating Hessian Matrix, we can use Fisher Information Matrix to simplify the process of calculating. 

Because the negative expected Hessian of log likelihood is equal to the Fisher Information Matrix 
F.

\begin{small}
	\begin{equation}
		\begin{aligned}
			F_{i j}=-\mathbb{E}\left[f^{\prime \prime}\left(\theta_{past}^{*}\right)\right]=-\mathbb{E}_{p\left(\theta \mid D_{past}\right)}\left[\left.\frac{\partial^{2} \log f(\theta)}{\partial \theta_{i} \theta_{j}}\right|_{\theta=\theta_{past}^{*}}\right]
		\end{aligned}
	\end{equation}
\end{small} where $\theta_i,\theta_j$ are elements of the $\theta^*_{past}$.

In order to reduce the amount of calculation, we can only get the diagonal elements.
\begin{small}
	\begin{equation}
		\begin{aligned}
			F_{i i}=-\mathbb{E}_{p\left(\theta \mid D_{past}\right)}\left[\left.\frac{\partial^{2} \log f(\theta)}{\partial \theta^2_{i}}\right|_{\theta=\theta_{past}^{*}}\right]
		\end{aligned}
	\end{equation}
\end{small}

According to the definition of the Fisher Information Matrix, we can know that

\begin{small}
	\begin{equation}
		\begin{aligned}
			F_{i i}&=\mathbb{E}_{p\left(\theta \mid D_{past}\right)}\left[\left(\left.\frac{\partial \log f(\theta)}{\partial \theta_{i}}\right|_{\theta=\theta_{past}^{*}}\right)^{2}\right]\\
			F_{j}&=\frac{1}{m} \sum_{i=1}^{m}\left(\frac{\partial \log f(\theta)}{\partial \theta_{j}}\right)^{2}
		\end{aligned}
	\end{equation}
\end{small}

Final step, we can approximate the $p d f$ with $f(\theta) \approx \mathcal{N}\left[\theta^*_{past}, \sigma^{2}\right]$.

\begin{small}
	\begin{equation}
		\begin{aligned}
			&f(\theta) \approx \mathcal{N}\left[\theta^*_{past}, \sigma^{2}\right] = \frac{1}{\sqrt{2 \pi} \sigma} e^{-\frac{(\theta-\theta^*_{past})^{2}}{2 \sigma^{2}}}\\
			&\log f(\theta) \approx \log \frac{1}{\sqrt{2 \pi} \sigma}-\frac{(\theta-\theta^*_{past})^{2}}{2 \sigma^{2}}
		\end{aligned}
	\end{equation}
\end{small}

According to equation (6)(9)(10)(11), we can get the loss function as follows.

\begin{small}
	\begin{equation}
		\begin{aligned}
			\theta_{i} &= \mathop{\arg\max}_{\theta_{i}} \log p(\theta_{i}|\mathcal{X},\mathcal{Y})\\
			&=\mathop{\arg\max}_{\theta_{i}}(\log p(\mathcal{X}_{i},\mathcal{Y}_{i}|\theta_{i}) -\log p(\mathcal{X}_{i},\mathcal{Y}_{i})+\log p(\theta_{i}|\mathcal{X}_{past},\mathcal{Y}_{past}))\\
			&= \mathop{\arg\max}_{\theta_{i}}(-\mathcal{L}+\log f(\theta_i))\\
			&= \mathop{\arg\max}_{\theta_{i}}(-\mathcal{L}+\frac{1}{2}(\theta_{i}-\theta_{past}^{*})^{2} (\frac{\partial^{2} f(\theta)}{\partial^{2} \theta}|_{\theta=\theta_{past}^{*}}))\\
			&= \mathop{\arg\min}_{\theta_{i}}(\mathcal{L}-\frac{1}{2}(\theta_{i}-\theta_{past}^{*})^{2} (\frac{\partial^{2} f(\theta)}{\partial^{2} \theta}|_{\theta=\theta_{past}^{*}}))\\
			&= \mathop{\arg\min}_{\theta_{i}}(\mathcal{L}+\frac{\lambda}{2} \sum_{i} F_{i}\left(\theta_{i}-\theta_{past, i}^{*}\right)^{2})
		\end{aligned}
	\end{equation}
\end{small}
where $\lambda$ is the another regularization and all constant values is ignored.

\clearpage
\newpage
{\small
	\bibliographystyle{ieee_fullname}
	\bibliography{ref}
}
\end{document}